\documentclass[journal]{IEEEtran}
\ifCLASSINFOpdf
\else
\fi

\usepackage{graphicx} 
\usepackage{epsfig} 
\usepackage{mathptmx} 
\usepackage{times} 
\usepackage{amsmath} 
\usepackage{amssymb}  
\usepackage{bm}
\usepackage{subfig}
\usepackage{url}
\usepackage[maxfloats=]{morefloats}
\usepackage{hyperref}
\usepackage{soul}
\usepackage{color}

\usepackage[ruled]{algorithm2e}

\SetCommentSty{mycommfont}

\usepackage{algpseudocode}
\algnewcommand\algorithmicforeach{\textbf{for each}}
\algdef{S}[FOR]{ForEach}[1]{\algorithmicforeach\ #1\ \algorithmicdo}

\DeclareMathOperator*{\argmin}{argmin}

\usepackage{lineno,hyperref}
\modulolinenumbers[5]

\begin{document}
%
\title{ Curved Buildings Reconstruction from Airborne LiDAR Data by Matching and Deforming Geometric Primitives}
%
%
%

\author{Jingwei~Song,~\IEEEmembership{Student member,~IEEE,}
		Shaobo~Xia,
        Jun~Wang,
        Dong~Chen$^*$,~\IEEEmembership{Member,~IEEE}
\thanks{This work was supported in part of National Natural Science Foundation of China under Grant 41971415. (Corresponding authors: Dong Chen)}
\thanks{Jingwei Song is with the Centre for Autonomous Systems, University of Technology, Sydney, P.O. Box 123, Broadway, NSW 2007, Australia. e-mail:(jingwei.song@student.uts.edu.au). }
\thanks{Shaobo Xia is with the Department of Geomatics Engineering, University of Calgary, T2N 1N4, Canada. e-mail: (shaobo.xia@ucalgary.ca). }
\thanks{Jun Wang is with the Institute of Remote Sensing and Digital Earth, Chinese Academy of Sciences, University of Chinese
	Academy of Sciences, Beijing, China. 
	e-mail:(wangjun@radi.ac.cn). }
\thanks{Dong Chen is with the College of Civil Engineering, Nanjing Forestry University, Nanjing 210037, China. e-mail: (chendong@njfu.edu.cn).}
}

%
%


\markboth{xxx}%
{Shell \MakeLowercase{\textit{et al.}}: Bare Demo of IEEEtran.cls for IEEE Journals}
%



\maketitle

\begin{abstract}
{Airborne} LiDAR (Light Detection and Ranging) data is widely applied in building reconstruction, with studies reporting success in typical buildings. However, the reconstruction of curved buildings remains an open research problem. To this end, we propose a new framework for curved building reconstruction via assembling and deforming geometric primitives. The input LiDAR point cloud are first converted into contours where individual buildings are identified. After recognizing geometric units (primitives) from building contours, we get initial models by matching basic geometric primitives to these primitives. To polish assembly models, we employ a warping field for model refinements. Specifically, an embedded deformation (ED) graph is constructed via downsampling the initial model. Then, the point-to-model displacements are minimized by adjusting node parameters in the ED graph based on our objective function. The presented framework is validated on several highly curved buildings collected by various LiDAR in different cities. The experimental results, as well as accuracy comparison, demonstrate the advantage and effectiveness of our method. {The new insight attributes to an efficient reconstruction manner.} Moreover, we prove that the primitive-based framework significantly reduces the data storage to 10-20 percent of classical mesh models.
\end{abstract}


%
\IEEEpeerreviewmaketitle

\section{INTRODUCTION}
\IEEEPARstart{A}utomatic 3D building reconstruction plays a more prominent role in Geography Information Systems (GIS), smart city, and environmental modelling applications compared to conventional 2D building footprints. Thus, the research of automatic reconstruction of urban 3D models {gained} {popularity} over the past two decades \cite{gruen2012automatic}, and is still an important topic in photogrammetric research \cite{haala2010update}. With the development of sensors, different types of data are utilized and reported, including airborne Light Detection And Ranging (LiDAR) \cite{song2015extraction}, airborne imagery \cite{daftry2015building},  oblique imagery \cite{rupnik2014oblique}, terrestrial laser scanning \cite{lin2013semantic}, RGB-D sensors \cite{loianno2015cooperative} and RGB sensors with Inertial Measurement Unit (IMU) \cite{lin2018autonomous}. Sensors acquiring RGB information are very sensitive to radiation levels and the effectiveness of the post-processing algorithm. Despite being able to collect data with a high level of detail, RGB-D sensors suffer greatly from small ranges and are only applicable for indoor or close-range photogrammetry. As {the authors} \cite{haala2010update} point out, LiDAR measurements are still the only feasible data source for automatic 3D building reconstruction in terms of accuracy, reliability, and detail. Airborne LiDAR directly generates a point cloud with a resolution ranging from centimetres to millimetres, and with a limited amount of noise and outliers. Therefore, airborne LiDAR is currently the most widely used sensor for large scale urban infrastructure reconstruction \cite{shan2018topographic}. Although there exist several hybrid methods to combine LiDAR with other data sources, such as images, the issue of how to optimally integrate data from the multiple data sources with dissimilar characteristics remains an open research problem. Recently, the improvements in LiDAR data quality provide an interesting opportunity to create 3D building models by using LiDAR points without the integration of other data sources. \par

{A general building reconstruction process includes building segmentation, feature recognition, and modelling {\cite{haala2010update}}. Building segmentation refers to the task of splitting a given point cloud into non-overlapping homogeneous parts that can be classified into several specific objects. After obtaining the raw point cloud, it goes through the feature recognition process, which is defined as the recognition of topological relationships between the segmented regions. With different types or characters of the building identified, {fine-scale building,} normally in mesh format, is reconstructed. The three sequential processes are technically consistent. According to {the research} {\cite{wang2018lidar}}, they are normally classified as data-driven, model-driven and hybrid approach.}\par

{Considerable efforts have been devoted in the data-driven approaches. Aiming at segmenting building from other objects, the data-driven studies achieve this in the manner of segmenting and reconstructing buildings directly from data without prior templates of buildings} \cite{rottensteiner2003automatic,miliaresis2007segmentation}. {The region growing method randomly specifies a seed point, and subsequently measures its similarity with the neighbours in order to determine a match. If successful, the new point will be incorporated into the region to search for more points. Edge-based methods search line features in the dataset to determine edge locations based on template building roofs. The connected edges constitute the surfaces of polyhedral building roofs, the main structures of typical polyhedral buildings. After the segmentation, the building is recognized with data-driven feature recognition steps. {The research} {\cite{elberink2009target}} defines adjacency relations that are dependent on the length of the segment intersection lines, determined by the points of both segments that are within a specific distance. Regions are considered as adjacent if the horizontal distance between two segments is smaller than a given threshold {\cite{park2006automatic}}.} {The authors} \cite{yan2014global}{ introduce a multi-label optimization strategy to achieve globally optimal building roof segmentation. Other approaches turn to edge labelling of orthogonality} \cite{verma20063d,shan2018topographic}, region symmetry \cite{milde2008building,huang2011rule} and rooftop or planar regularity constraint \cite{jung2017implicit}\cite{hou2018planarity}. {Overall, the definition of topological relationships is highly dependent on building types and, in turn, greatly impacts modelling. The modelling process for model-driven approach involves connecting corners as the intersections and connections of extended lines and step edges, and can be categorized as the simple primitive matching and combined primitive modelling.}\par

{In addition to the data-driven approach, the model-driven approach is also popular in the state-of-art building reconstructions {\cite{sampath2010segmentation}}. Model-driven tackles the segmentation by employing the mathematical expressions of the target object to search buildings {\cite{sohn2008using}}. Among them, the Hough transform is a popular approach in building recognition {\cite{maas1999two}} {\cite{vosselman20013d}} {\cite{hulik2014continuous}}. It simultaneously searches for the planes, as well as the corresponding positions and orientations. RANdom SAmple Consensus (RANSAC) is also adopted for building segmentation. It randomly and iteratively samples a small subset of points to determine plane parameters, which in turn are tested against input point cloud. The algorithm stops when the Euclidean difference error is small enough. The aforementioned segmentation approaches have limitations, such as poor robustness and sensitivity to noise. To address the problem, recent progresses including B-spline fitting {\cite{dimitrov2016non}}, Gestalt laws {\cite{hu2018towards}}, structural regularity {\cite{chen2018automatic}}, regular object-based {\cite{li2019geometric}} and complex 3D family-based model fitting {\cite{xue2018automatic}} are proposed and reported to have better performance. When the model is segmented, the feature recognition step, specifically the rooftop primitives recognition, determines the common borders with a predefined roof types. This is beneficial, particularly in polyhedral building roofs, where strict topological relationships among planes and edges exist. Different methods define various topological relationships. {Recent research} {\cite{chen2017unsupervised}} proposes a method to recognize the volumetric building structure. In general, the extracted features in feature recognition step is utilized to parameterize modelling. The data-driven based 3D modelling process is carried out with prismatic modelling, polyhedral modelling {\cite{yan2015automatic}} {\cite{yang2017automated}} and nonlinear rooftop modelling {\cite{zhou2008fast}}. Based on the extracted features, rooftop primitives is enforced and 3D model is reconstructed.}\par

{Aiming at taking advantage of both model-driven and data-driven, recent researches propose hybrid approaches to improve the accuracy and robustness} \cite{tian2010knowledge,arefi2013building,xu2015investigation,song2015extraction}. {Specifically, to tackle the complex buildings, {\cite{lafarge2012creating}}, {\cite{vetrivel2015identification}} and {\cite{nan2017polyfit}} present solutions on more complete models of unspecified urban environments and contribute to the hybrid reconstruction including 3D geometric primitives or polygonal surface. These primitives are then used to represent standard roof sections and mesh-patches to describe more irregular roof components. Although other methods reconstruct the irregular rooftops via the meshed DSM (digital surface model),  the created model is just a giant mesh, thereby losing the structural information.}\par 

After {the} tremendous efforts, 3D regular polyhedral building reconstruction with planar roofs and facades has been achieved with great progress, and can be deployed widely for practical applications. The base for conventional polyhedral building reconstruction lies in {the} key planar, edge, and corner recognition and extraction. {The continuous curved buildings, alternatively, cannot be formulated with the discontinuous manner in polyhedral buildings. The successful classical polyhedral algorithms cannot be directly applied in 3D curved building reconstruction because of the free form definition in corner, curves or surfaces.}  Aiming at processing the curved buildings, our previous work \cite{song2015extraction} presents a contour clustering approach to segment and reconstruct the buildings with the curved topology, where the contour lines are clustered and derived. The rooftop type is then identified by analysing the enclosed topological relationship between the contour lines. Following this, the rooftop is reconstructed by fitting the corresponding separated LiDAR points. The model is reconstructed in the form of meshes. This approach successfully overcomes conventional line and corner extraction limitations and proves to be feasible for the application of curved features. Following this work, {researchers} \cite{gao2017automatic} and \cite{gilani2018segmentation} introduce {achieve positive results in complex building reconstruction.} \cite{he20163d} makes full use of contour shape clustering in damage building extraction, where the shape is highly irregular. {The work} \cite{wu2017graph} implements the same framework \cite{song2015extraction} and denotes it as the graph-based method. \par  

However, {these approaches are inconsistent to conventional polyhedral building and the free form shape poses heavy storage burden to the systems.} {For all previous researches  {\cite{song2015extraction}\cite{he20163d}\cite{wu2017graph}}, no work pay attention to the disadvantages of the free form mesh models, including the inconsistency to standard template-based models and large memory occupation.} A polyhedral model only needs corners and their topological relationships. A mesh expressing curved surface or roof, however, requires a large sum of patches, far more than a regular building. Inspired by \cite{tulsiani2017learning}, which uses generalized cylinders to represent the structure of any shape, {and also aiming at in consistent with primitive approach in polyhedral building recovery,} we propose the modelling of coarse buildings with simple geometric primitives, such as a hemisphere, cone, cylinder, and polyhedron, and employ a warping field to deform and polish the coarse reconstruction. Distinct from \cite{lafarge2012creating} and \cite{vetrivel2015identification}, {who uses the simple geometries}, our model deforms itself to express a more complex scenario. A warping field is employed to learn the deformation from the simple geometry to a point cloud. Correspondingly, we substitute the conventional contour clustering {approach {\cite{song2015extraction}} with a modified Procrustean analysis method} for dividing the building into primitives. We theoretically prove that the reconstructed model significantly reduces storage whilst maintaining sufficient detail. Overall, there are mainly {three} contributions in this paper.\par
\begin{itemize}
    \item Comparing with mesh-based {curved} modeling, we present an innovative complete framework for curved surface building reconstruction based on geometric primitives and their deformations. {It is consistent with conventional primitive based polyhedral modelling.}
    \item  {A modified Procrustean} analysis approach is proposed for dividing the building into primitives for geometric primitive matching. 
    \item Deformation field is introduced to minimize the differences between models and point cloud for the first time.
\end{itemize} 
 {The rest of the paper} is presented as follows. Section 2 describes the details of the complete framework, from segmentation, contour clustering, primitive recognition and the formulations of the deformation field. Section 3 demonstrates the resulting geometric primitive based models and comparisons with previous {mesh-based} models \cite{song2015extraction}. Lastly, Section 4 concludes this research.\par

\section{Methodology}
\begin{figure*}[]
	\centering
	\includegraphics[width=1\textwidth]{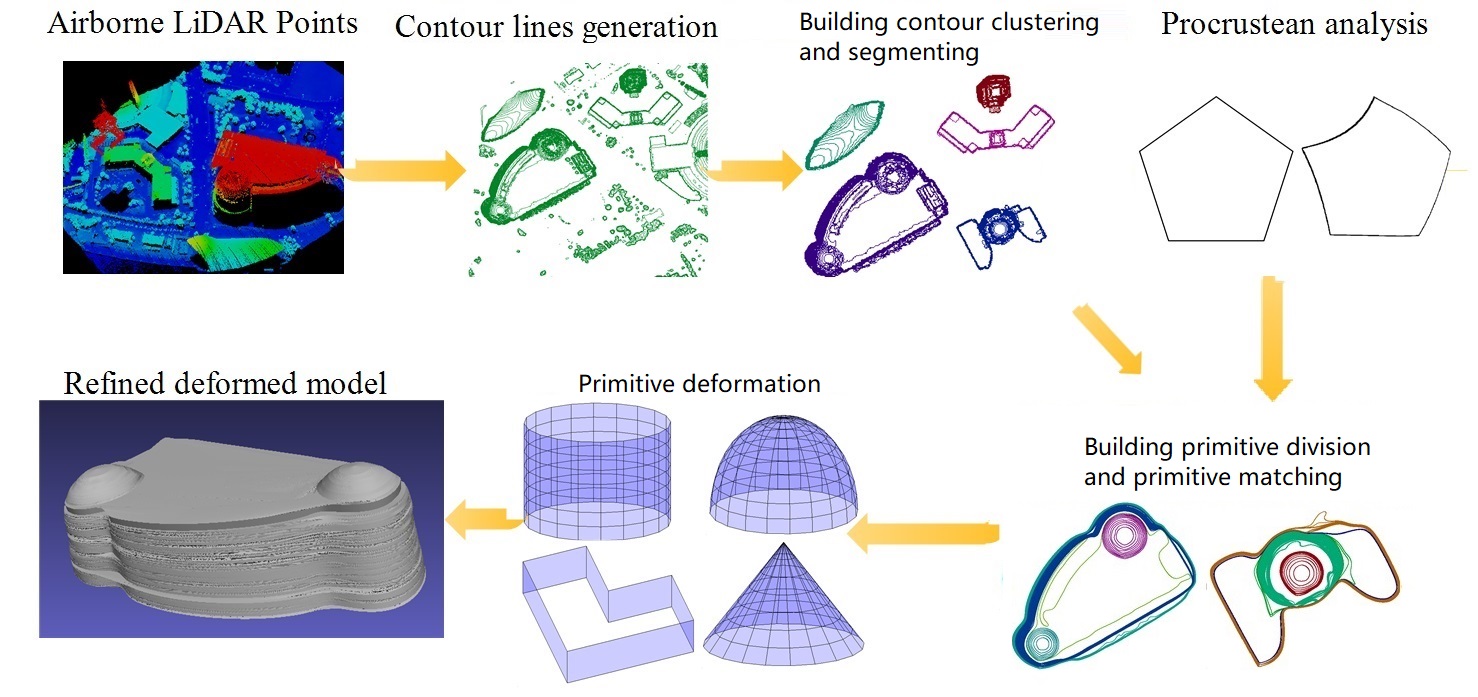}
	\caption{The framework of our curved building extraction and reconstruction.}
	\label{fig:pipeline}
\end{figure*}
\subsection{Overview}
Our previous research \cite{song2015extraction} {provides the framework with free form curved building extraction and reconstruction. In this research}, {we adopt and improve the contour-based building extraction method,} and propose a noval {primitive based} curved building modelling framework. The complete system consists of contour generation, building segmentation, contour clustering, building primitive division, primitive matching, and model refinements by deformation. Fig. \ref{fig:pipeline} {presents} the pipeline of the proposed approach. 
\par We adopt the contour line extraction and building segmentation method used in \cite{song2015extraction}. Contours are directly generated from raw LiDAR point cloud data with morphological filter and building point identification \cite{mccullagh1988terrain}. Following this, an inner-nest topological relationship is exerted on the contour lines in order to cluster the contours into individual clusters. We select properties such as area, height, and circularity of building contours to automatically identify individual buildings, based on a previous polygon center topology algorithm \cite{song2015extraction,gao2017automatic}. Correspondingly, the associated point cloud is segmented from the dataset according to the clustering of contour lines. Furthermore, each building is divided into {metamodels (termed as primitives)} and fitted with {the} predefined geometric primitives. We employ an improved {Procrustean} analysis algorithm for primitive division. All primitives are assembled to reconstruct the building. Finally, in order to make the reconstructed curved surface more vivid and accurate, we introduce a deformation field to deform the coarse primitives, refining it to be very close to the target point cloud. The final reconstructed models are expressed in the form of parameterized primitives, their positions and orientations, and associated deformation graphs.
\subsection{Model format: Geometric primitive based representation}
\label{subsection_modelformat}
The key concept in our paper is the assembly of buildings by composing the predicted simple transformed and deformed primitives. These primitives include simple geometries, such as hemispheres, cones, cylinders, and polyhedrons. After the complete building is segmented into small primitives, each model (in point cloud format) is split into small primitives, and are fitted with predefined primitives. Therefore, we {propose} that each component stores four domains: {primitive type and associated parameters in canonical space $\mathbf{\zeta}_i$, primitive rotation $\mathbf{R}_i  \in SO(3)$ and primitive translation $\mathbf{T}_i  \in \mathbb{R}^3$ for the transformation of primitive from canonical space to target position.}\par 

\begin{figure}[]
	\centering
	\includegraphics[width=0.5\textwidth]{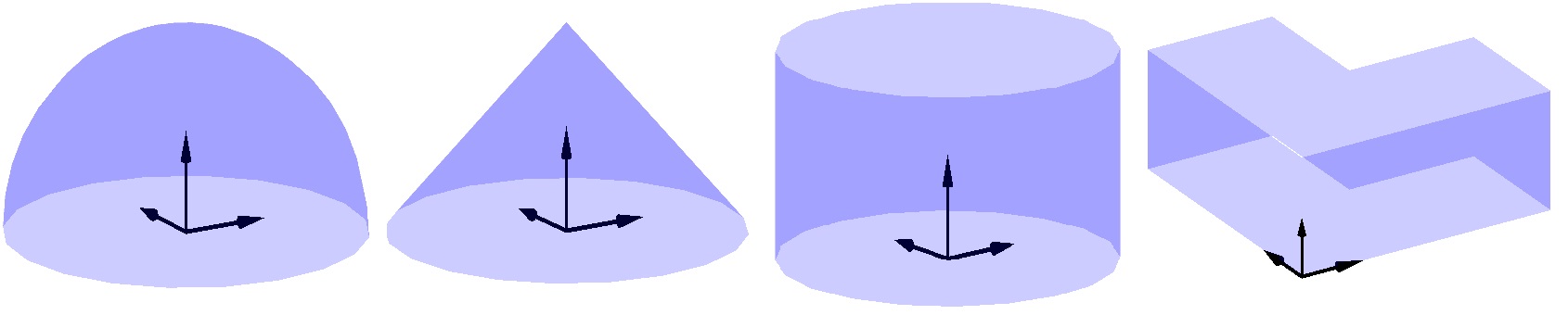}
	\caption{{The definition of the canonical coordinate system of the primitives.} Note that the last is the {polyhedral primitive} and with flexible number of corners and edges.}
	\label{fig:primitive_cord}
\end{figure}

Fig. \ref{fig:primitive_cord} shows the coordinate definitions of {the} primitives. Hemisphere is parameterized with the centre and the radius. Cone is defined by the bottom circle (centre and radius) and the facade in the form of $z=a(x^2+y^2)+b$, where $x,y,z$ are the coordinates and $a,b$ the coefficients. The cylinder is defined by the bottom circle (centre and radius) and the height. The polyhedron is defined by a chain of points at the bottom and a corresponding chain of points at the top. Note that the facade of the polyhedral does not necessarily need to be strictly vertical. The local coordinate centres for the hemisphere, cone, and cylinder are defined at the centre of the bottom circle. For the polyhedral, it is defined as a random corner of the bottom polygon. {Note that the bottom and top of polyhedral primitive are the same in our definition.} Here, the Z coordinate always points upward. \par 

Although we attempt to model with `pure’ primitives such as cones, hemispheres or cuboids, {primitive with higher degree of freedom} (polyhedron) is present at a low degree during fitting like the model shown in the last model of Fig. \ref{fig:diagram_cluster}. In addition, complex buildings are composed of convex and even concave structures. It is much cheaper to encode them with one polyhedron rather than multiple cuboids. Therefore, we introduce the coarse polyhedron (with flexible corners and edges) as a primitive in this framework, since the corners of a polyhedron are sparse, and thus the memory cost remains small.\par

\subsection{Building contour clustering and segmenting}
\label{subsection_Segmentation_Clustering}
After acquiring LiDAR point cloud, we generate contour lines and segment buildings contours from other objects. Each building is extracted in the form of contour clusters. {Before moving to the primitive based fitting}, we first review building segmentation from contour clusters. A contour is defined as a line of all points sharing the same height. In the geography community, a contour map is a powerful tool used to express 3D values (height, temperature, pressure, and moisture) in a 2D format. It accurately preserves the original spatial distribution of 3D values and keeps the topology of this variation. Due to the inherent character of contour lines, they greatly help to overcome the problem of the inability of the curved building to be expressed using corners and lines. Therefore, we still adopt the contour clustering and segmentation approaches described in \cite{song2015extraction}. The major difference in this paper is that we add a morphological filter \cite{chen2014methodology} to do a coarse filter of ground points. By introducing a morphological filter, the quality of {the} generated contours are greatly improved, benefiting subsequent contour based building division. Although the topology of {the} contours can filter most non-building points, the contour itself is susceptible to neighboring non-building points, and thus a filtering approach is necessary to enhance robustness.\par

 Given a raw LiDAR point cloud, after filtered with morphological filter \cite{chen2014methodology}, contours are created by traversing a set of points with a neighbouring surface of the same elevation. All points sharing the same elevation are connected to form a polyline. After the generation of contours from the LiDAR point cloud, open polygons are removed to avoid reconstructing buildings partially observed. An {inner-nested} topology constraint is then {carried out on} the contours to segment the contour lines into clusters. In order to process concave polygons, we substitute the centre estimation algorithm \cite{song2015extraction} with a fast general polygon centroid estimation method \cite{bourke1988calculating}. \par 
%
{For} the non-intersect property of contours \cite{antenucci1991geographic}, two specific polygons are either inter-nested or separated. Hence, we {utilize} the centre of the contours to test if one contour is nested in another contour, and cluster all contour lines into individual {for primitive fitting}. As described in \cite{song2015extraction}, common objects such as terrain, individual trees, small area trees, forest, power lines, and automobiles, are different in terms of area, height and circularity. Based on these attributes, we segment clusters of the building using the recommended thresholds presented in work \cite{song2015extraction}. After this step, we have each building in the form of contour clusters. For details on contour topology based segmentation, please refer to \cite{song2015extraction} for technical details.\par

\subsection{Building primitive division and primitive matching}
\label{building_division}
\begin{figure}[]
	\centering
	\includegraphics[width=0.3\textwidth]{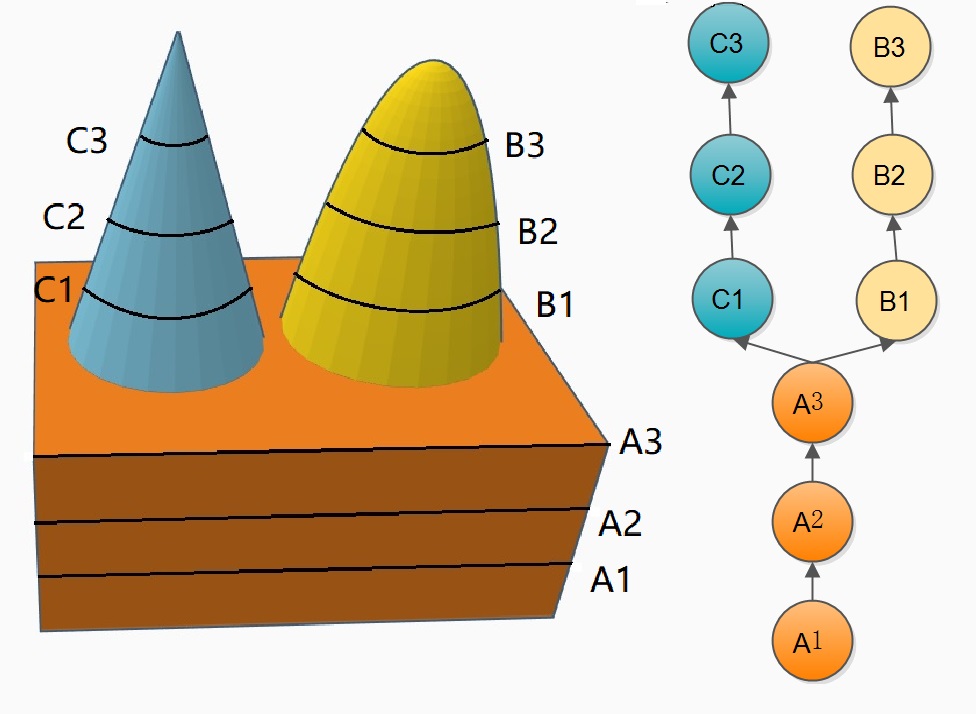}
	\caption{Typical building contours and its cluster trees.}
	\label{fig:diagram_cluster}
\end{figure}
\begin{figure}[]
	\centering
	\includegraphics[width=0.3\textwidth]{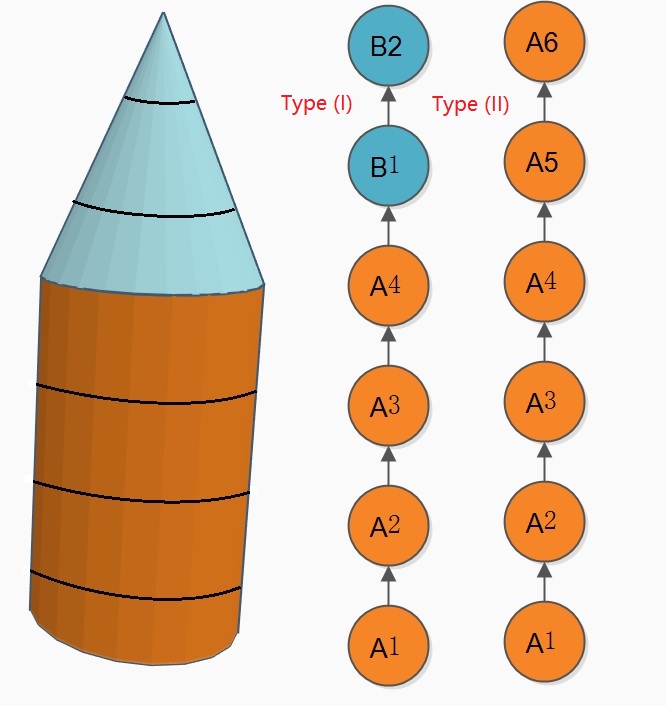}
	\caption{Comparisons of the contour cluster division. {It shows two types of building segmentation.} Type (I) is the building contour division results {of MPA method proposed in} this paper. Type (II) is the results in \cite{song2015extraction}.}
	\label{fig:diagram_cluster2}
\end{figure}
Now, we turn specifically to dividing complete building into contour clusters and match with primitives. {Previous researches \cite{song2015extraction}\cite{wu2017graph} model the 3D building in a straightforward manner, that is the free form mesh model. This method is easy to be implemented but is inconsistent with other polyhedral modelling and poses heavy storage burden. This section adopts the geometric primitive to roughly fit the curved building.} Traditional planar roof face detection is highly suited to conventional multiple planar building reconstruction. For complex buildings with curved surfaces, however, planar roofs fail to describe curved edges and surfaces. \cite{song2015extraction} adopts contour clusters to split buildings into several independent roofs. These roofs are matched with predefined roof types based on the changes of areas as the height increases. {As presented in Fig. {\ref{fig:diagram_cluster}}, the building contours are segmented based on the topological relations among neighboring contours in Z dimension. The same inner-nested contour cluster is categorized as a unit, with the corresponding point cloud extracted for roof fitting. Similar researches} \cite{he20163d} and \cite{wu2017graph} adopt this {strategy} and prove its feasibility. {As demonstrated in Fig. {\ref{fig:diagram_cluster}}, this process works well because the 3 divided components are simple units and can be modelled in mesh format However, if complex or hybrid geometric primitives (see Fig. {\ref{fig:primitive_cord}}) are introduced for component recognition, a more specific method should be incorporated for finer primitive segmentation. Fig. {\ref{fig:diagram_cluster2}} is a counter-example, the work of {\cite{he20163d}} and {\cite{wu2017graph}} cluster the building (Fig. {\ref{fig:diagram_cluster2}}) into one primitive (Type II), but this example {is} an assembly of two primitives (cylinder and cone, shown in Type I). Conventional approaches cluster the building contour into one whole model, which works fine for mesh based buildings. However, this model cannot be fitted with one single primitive. An improved method is required to separate the model into 2 small primitives, fitted with primitives independently.}\par 

In addition to {the} inner-nested topology, a more accurate {strategy is to} compare the similarity of one contour with its upper and lower neighbours. Fig. \ref{fig:diagram_cluster2} shows that contours within a primitive are similar in shape but may have different scales, rotation, and translation. Thus a similarity measurement is required for clustering building contours into small primitives. Contours possess attributes including area, length, the circular index, and the chaos index \cite{he20163d}. However, we note that these 0-dimensional indices are not able to measure 2D similarity. Therefore, this paper proposes a modified {Procrustean} analysis (MPA) method for defining and measuring similarity within contours. \par

{The Procrustean} analysis (PA) \cite{gower1975generalized} is a method used to align a set of curvatures close to each other in shape. In Euclidean space, PA searches the optimal orthogonal transformation (rotation, translation, and scaling) to align one shape to another. After obtaining the best transformation, the {Procrustean} distance is defined by the average distance from the transformed shape to the target shape. Similarly, the {Procrustean} space is presented for all shapes. {Provided with} two matrices $\mathbf{P}$ and $\tilde{\mathbf{P}}$ containing {sets of 2D points}, e.g. $n$ points {in $\mathbb{R}^2$}. The goal is to optimize the rotation matrix $\mathbf{R}\in SO(2)$, translation vector $\mathbf{t}\in \mathbb{R}^2$ and scale scalar $c$, by minimizing the distance defined as:

\begin{equation}
\begin{split}
\argmin\limits_{\mathbf{R},\mathbf{t},c} 
\sum_{i=1}^n||\tilde{\mathbf{p}}_i-c\mathbf{R}\mathbf{p}_i-\mathbf{t}||^2,
\end{split}
\label{energyprocrustean_1}
\end{equation}
where $\tilde{\mathbf{p}}_i$ and $\mathbf{p}_i$ are the $i$th point of $\mathbf{P}$ and $\tilde{\mathbf{P}}$.

Conventional PA formulation considers 4 degrees of freedom, including rigid transformation and scaling. In contour line matching, however, shear transformations, such as geometric distortions and deformations, are presented. In practice, some curved surfaces are oblique, or even more complex (refer to experiments like those in Fig. \ref{fig:reconstruction_sandiego} and Fig. \ref{fig:reconstruction_compare_withOPTIK}), for only rigid and scaling alignment.  Therefore, we add more degrees of freedom by relaxing the rotation matrix {$\mathbf{R} \in SO(2)$} to affine matrix {$\mathbf{S} \in \mathbb{R}^2$}. In other words, $\mathbf{S}$ is a $2 \times 2$ affine matrix {with 4 degree of freedom instead of 1 in PA}. Derived from {the Procrustean} analysis, the proposed MPA covers all situations relating to the topological relations or {shear transformation} within different contours. The proposed MPA formulation is defined here as follows:\par

\begin{equation}
\begin{split}
\argmin\limits_{\mathbf{R},\mathbf{t},c} 
\sum_{i=1}^n||\tilde{\mathbf{p}}_i-\mathbf{S}\mathbf{p}_i-\mathbf{t}||^2,
\end{split}
\label{energyprocrustean_2}
\end{equation}
With its linearity, the energy function in Eq. (\ref{energyprocrustean_2}) converges much faster than Eq. (\ref{energyprocrustean_1}). Similarly, the MPA distance $\mathrm{\tilde{d}}$ is defined as the following:

\begin{equation}
\mathrm{\tilde{d}} = \mathrm{d}/\mathrm{a},
\end{equation}
where $\mathrm{d}$ is the average {the Procrustean} distance {defined as the objective function Eq. (\ref{energyprocrustean_1}), which measures the point-wise average distance} from affine transformed $\mathbf{P}$ to $\tilde{\mathbf{P}}$. $\mathrm{a}$ is the area of $\tilde{\mathbf{P}}$. {In the conventional formulation, the nonlinear rotation matrix in Eq. ({\ref{energyprocrustean_1}}) requires an iterative process to solve the equation. The advantage of the proposed linear modified formulation Eq. ({\ref{energyprocrustean_1}}) is that it yields closed-form solution and the time consumption is significantly {reduced}.}\par 

{With the definition of} the {Procrustean} distance, each contour {is} compared with its upper and lower neighbouring contour to test {the} similarity. Contours that are close in {the Procrustean} space are clustered {as} primitives. Correspondingly, this process divides contours and associated point cloud into units of building for primitive model fitting. Each primitive contains several similar contours for fitting primitive geometries.\par

Primitive model fitting is achieved using a {2D} distance field function. Each grid in the uniform function records the distance to the nearest surface of a primitive by traversing all contour vertices. To simplify the initialization step, we {arbitrarily} define the pose of primitives within the volume. We first try to fit the parameterized cone, hemisphere and cylinder to the point cloud. If the matching error is no larger than the threshold, the {polyhedral primitive} is used for modelling. The parameterized cone, hemisphere and cylinder (parameter space $\mathbf{\zeta}_i$) are discretized into numbers of corners $\mathbb{V}(\mathbf{\zeta}_i)$ in canonical space.  The point cloud ($\mathbb{P}$) is used to build a volume based distance function which records the distance of each voxel to its nearest point $\mathbf{p}_i \in \mathbb{P}$.  We denote $\mathcal{F}(\cdot)$ as the field function recording the distance value on a specific grid.  A sufficient objective function ensures that the distance field of the assembled shape tends to zero for all points on the transformed discrete model:\par

\begin{equation}
\argmin\limits_{\mathbf{R},\mathbf{t},c} 
||\mathcal{F}(\mathbf{R}_i\mathbb{V}(\mathbf{\zeta}_i)+\mathbf{T}_i)||^2,
\end{equation}

\noindent {where the $\mathbf{R}_i$ and $\mathbf{T}_i$ are the rigid rotation and translation.} A threshold is applied to test the average error from points to surfaces of cone, hemisphere or cylinder (First three models shown in Fig. \ref{fig:primitive_cord}). If not obtained, the last primitive which is a polyhedron model is employed and the process degenerates into a coarse mesh based reconstruction. As claimed in Section \ref{subsection_modelformat}, there are many complex models which cannot be fitted with `simple' {primitives} like cone, hemisphere, and cylinder; {polyhedral primitive} is needed for comparisons.\par

\subsection{Model refinement with deformation field}
\label{model_EDfield}

\begin{figure}[]
	\centering
	\includegraphics[width=0.5\textwidth]{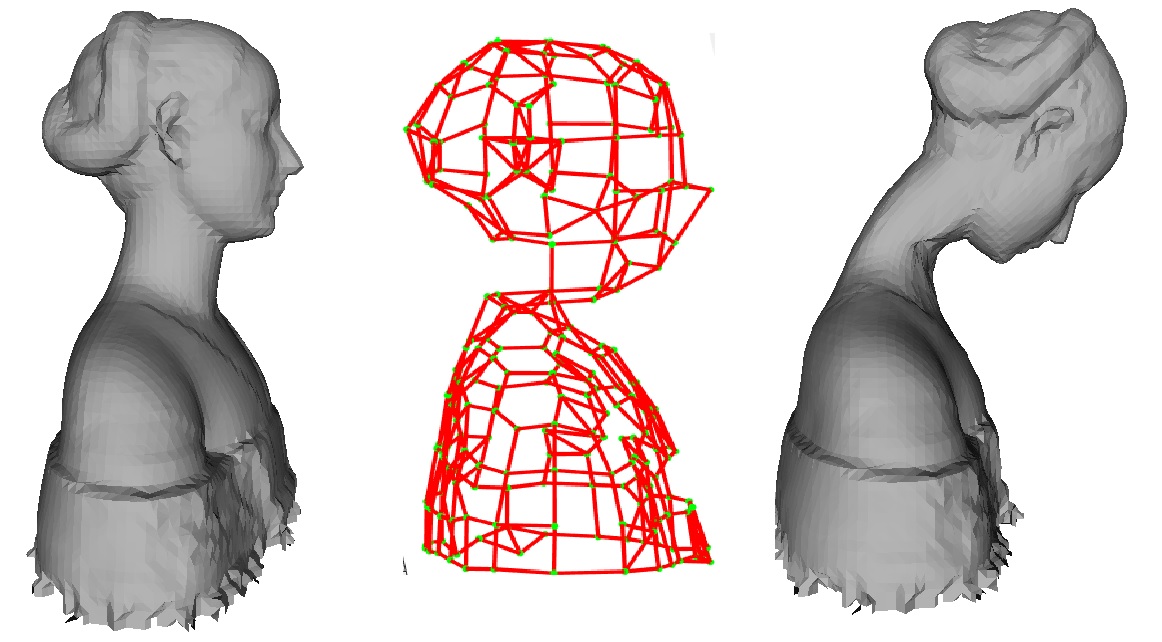}
	\caption{The example of ED based deformation. The model on the left is the original mesh model. The ED graph is presented in the middle. Green points represent the ED nodes, while red lines are the connecting edges. The deformed model is shown on the right, with head bending downwards.}
	\label{fig:ED_demo}
\end{figure}

{Section \ref{building_division} substitutes conventional free form curved building modelling with geometric primitive fitting, to solve the inconsistency of polyhedral modelling and heavy storage burden. However, the coarse fitting cannot provide fine scale curved building reconstruction. Therefore, to} further refine the reconstruction models, a warping field is introduced to polish the reconstructed primitive model assembly. Although geometric primitives provide basic structures and the coarse surface of buildings, it is unable to present detailed surfaces. Inspired by the achievements of human body reconstruction in the computer vision community \cite{newcombe2015dynamicfusion, dou2016fusion4d, innmann2016volumedeform}, we deform the coarse reconstruction to fine models. An embedded deformation (ED) \cite{sumner2007embedded} graph is employed as the warping field. This algorithm has been widely used in the field of non-rigid reconstruction \cite{newcombe2015dynamicfusion,whelan2016elasticfusion,innmann2016volumedeform}. Fig. \ref{fig:ED_demo} shows a typical example of ED based deformation. The deformation field consists of sparse nodes, with each node defining a local rigid rotation and transformation. The nodes are generated by downsampling original model (vertices) into sparse points. Downsampling is achieved by using uniform voxel grid to segment original vertices and average points in each grid. Each vertex is transformed to the target position by several nearest ED nodes, defined by position $\mathbf{b}_j$ $\in\mathbb{R}^3$, affine matrix $\mathbf{A}_j$ $\in\mathbb{R}^{3\times3}$ and translation vector in $\mathbf{t}_j$ $\in\mathbb{R}^3$. In practice, nodes are downsampled from the model for describing deformation; but it should be emphasized that nodes only define deformation and are irrelevant of the model. After downsampling, the node parameters $\mathbf{A}_j$ and $\mathbf{t}_j$ are initialized with the identity matrix and zero vector, respectively. For any given vertex $\mathbf{v}_i$, deformed position $\mathbf{\tilde{v}_i}$ is defined by the ED nodes as:\par 
\begin{equation}
\hat{\mathbf{v}}_i=\sum_{j=1}^k w_j(\mathbf{v}_j)[\mathbf{A}_j(\mathbf{v}_j-\mathbf{b}_j)+\mathbf{b}_j+\mathbf{t}_j],
\label{TransformationFomulation}
\end{equation}
where there are $k$ neighbors. In this research, $k$ is set to 4. The parameter $w_j(\mathbf{v}_j)$ is the quantified weight used to transform $\mathbf{v}_j$, applied by each related ED node. The weight is defined in the following form:

\begin{equation}
\label{eq_weight}
w_j(\mathbf{v}_j)=(1-||\mathbf{v}_j-\mathbf{b}_j||/d_{max}),
\end{equation}
where $d_{max}$ is the maximum distance of the vertex to the $k + 1$ nearest ED node. \par 

\begin{figure*}[!h]
	\centering
	\subfloat[Scene I]{
		\begin{minipage}[]{0.31\textwidth}
			\centering
			\includegraphics[width=1\linewidth]{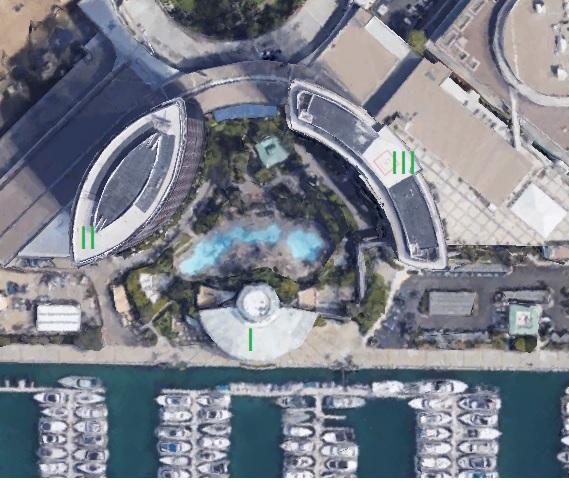}
		\end{minipage}
	}\/
	\subfloat[Scene II]{
		\begin{minipage}[]{0.31\textwidth}
			\centering
			\includegraphics[width=1\linewidth]{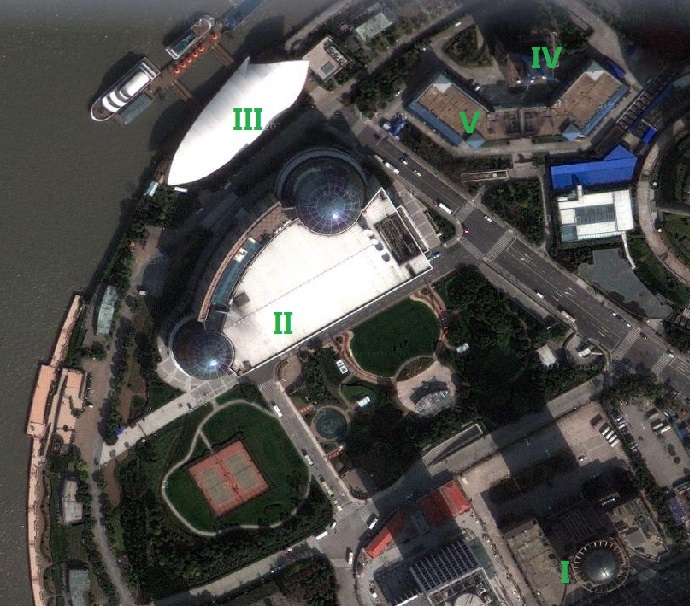}
		\end{minipage}
	}\/
	\subfloat[Scene III]{
		\begin{minipage}[]{0.31\textwidth}
			\centering
			\includegraphics[width=1\linewidth]{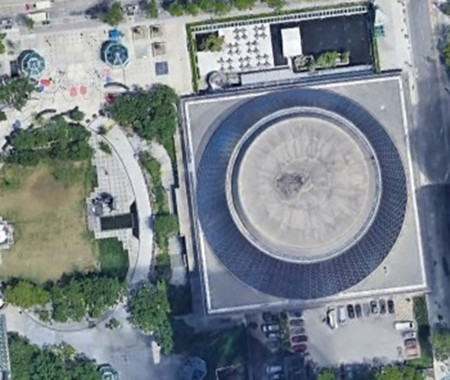}
		\end{minipage}
	}\/
	\subfloat[Scene I]{
		\begin{minipage}[]{0.31\textwidth}
			\centering
			\includegraphics[width=1\linewidth]{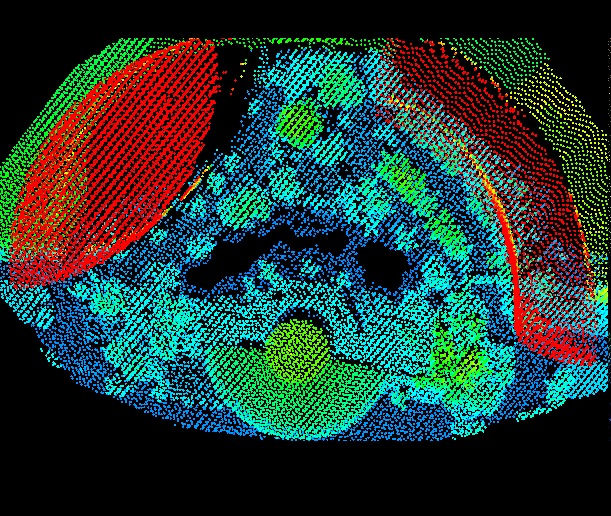}
		\end{minipage}
	}\/
	\subfloat[Scene II]{
		\begin{minipage}[]{0.31\textwidth}
			\centering
			\includegraphics[width=1\linewidth]{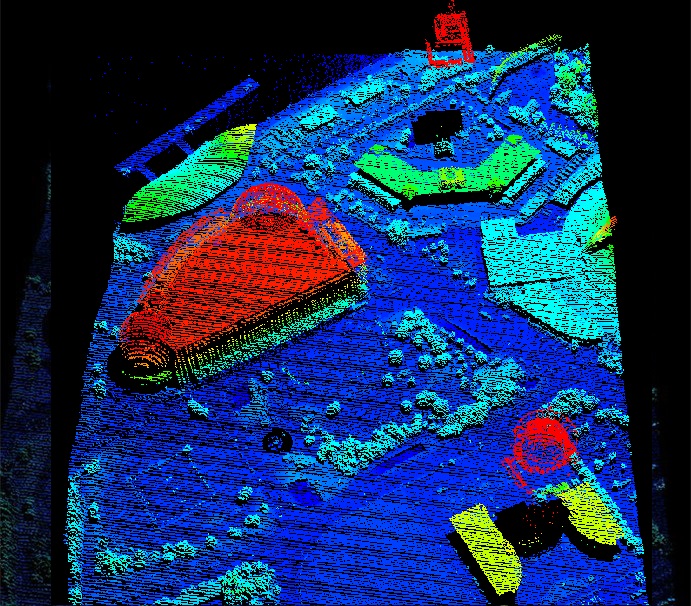}
		\end{minipage}
	}\/
	\subfloat[Scene III]{
		\begin{minipage}[]{0.31\textwidth}
			\centering
			\includegraphics[width=1\linewidth]{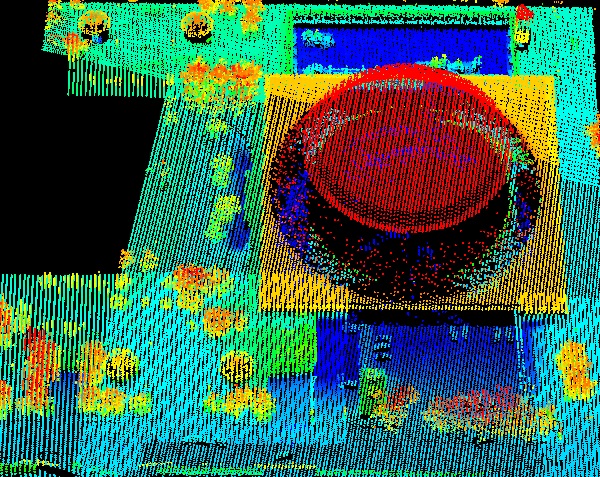}
		\end{minipage}
	}\/
	\caption{Three datasets tested in this paper. (a), (b) and (c) are aerial images and (d), (e) and (f) are LiDAR point cloud.}
	\label{fig:3datasets}
\end{figure*}

With the recognition of a primitive, we set up a warping field for each primitive to polish the curved surfaces. The objective function is composed of 3 terms, rotation, regularization, and the point-to-point correspondence distances:\par
\begin{equation}
\begin{split}
\argmin\limits_{{\mathbf{A}_1},{\mathbf{t}_1}...\mathbf{A}_m,\mathbf{t}_m} &w_{rot}E_{rot}+w_{reg} E_{reg}+w_{data} E_{data},
\end{split}
\label{energyfunction}
\end{equation}
where $m$ is the number of ED nodes.  Similar to \cite{sumner2007embedded}, \textbf{Rotation} and \textbf{Regularization}  rotation and regularization is used to avoid unreasonable deformations.\par 

\noindent\textbf{Rotation}. Defined in the form of an affine matrix, $\mathbf{A}_j$ is a relaxed rotation matrix. Orthogonality is imposed on $E_{rot}$ with the sums of the rotation error of all the matrices in the following form:
\begin{equation}
E_{rot}=\sum_{j=1}^m Rot(\mathbf{A}_j),
\end{equation}
\begin{equation}
\begin{aligned}
Rot(\mathbf{A}_j)=(\mathbf{c}_1\cdot\mathbf{c}_2)^2+(\mathbf{c}_1\cdot\mathbf{c}_3)^2+(\mathbf{c}_2\cdot\mathbf{c}_3)^2+\\
(\mathbf{c}_1\cdot\mathbf{c}_1-1)^2+(\mathbf{c}_2\cdot\mathbf{c}_2-1)^2+(\mathbf{c}_3\cdot\mathbf{c}_3-1)^2,
\end{aligned}
\end{equation}
where $\mathbf{c}_1$, $\mathbf{c}_2$ and $\mathbf{c}_3$ are the column vectors of the affine matrix $\mathbf{A}_j$.

\noindent\textbf{Regularization}. To prevent divergence of the neighbouring nodes exerting to the overlapping space, the regularization term is exerted on the objective function: \par
\begin{equation}
E_{reg}=\sum_{i=1}^m\sum_{j=1}^k\alpha_{ij}||\mathbf{A}_j(\mathbf{g}_i-\mathbf{b}_j)+\mathbf{b}_j+\mathbf{t}_j-(\mathbf{g}_i+\mathbf{t}_i)||^2,
\end{equation}
where $\alpha_{ij}$ is the weight calculated from the Euclidean distance between two ED nodes. We follow \cite{sumner2007embedded} by uniformly setting $\alpha_{ij}$ to 1.  \par

\noindent\textbf{Correspondence}. 
In section \ref{building_division}, $\mathcal{F}(\cdot)$ is used to define the energy function in the space measuring plane-to-point distances of the transformed primitive and point cloud. However, this energy function may not be fully satisfied because of primitive rigidity.  Therefore, we further utilize the energy function $\mathcal{F}(\cdot)$ and further deformed primitive to fit the point cloud. Minimizing this distance is equivalent to deforming the geometric primitive very close to the observations. Given the discretized primitive vertices $v_i \in \mathbb{V}(\mathbf{\zeta}_i)$, we have:

\begin{equation}
E_{data}=\sum_{v_i \in \mathbb{V}(\mathbf{\zeta}_i)}||\mathcal{F}(\mathbf{w}_j(\mathbf{v}_j)[\mathbf{A}_j(\mathbf{v}_j-\mathbf{b}_j)+\mathbf{b}_j+\mathbf{t}_j])||^2,
\label{DFF_equation}
\end{equation}

The Levenberg-Marquardt (LM) algorithm is utilized to solve the non-linear optimization problem. In addition to the conventional Gauss-Newton (GN) optimization method, LM adopts an extra term $\mu\mathbf{I}$ controlling the aggressiveness of the degrading. The damping factor $\mu$ can be manually adjusted to control the speed of convergence.\par

\section{Results and discussion}
\subsection{Synthetic data and in-situ dataset}
We present one synthetic and three real world experiments to demonstrate the effectiveness of the modified {Procrustean} analysis building division, memory usage of the model and the accuracy of fitting with a deformation graph. It is important to point out that in this work, we concentrate on primitive based building reconstruction given segmented building contour clusters (contour clusters of each building). We do not intend to address building contour cluster segmentation, although this is improved with some primitive strategies as shown in Fig. \ref{fig:diagram_cluster2}. In this section, we start with building primitive extraction by {the Procrustean} analysis based on each extracted building cluster.\par

To validate the feasibility of the proposed method, we extensively test it on {5} datasets and compare with previous work. The first airborne dataset (Scene I shown in Fig. \ref{fig:3datasets} (a) and (d)) was acquired over the San Diego city centre, US, from 16 March 2005 to 12 May 2005, and was provided by the US Geological Survey \footnote{http://opentopo.sdsc.edu/lidarDataset}. The median point density is 1.41 $pts/m^2$. Three typical curved buildings are chosen for automatic reconstruction. Two buildings in Fig. \ref{fig:3datasets} (II and III) are free form buildings, while the remaining building (I) is typically primitive based.\par 

For comparison with previous mesh based work, a second test is performed on the dataset of \cite{song2015extraction}, which was collected over Lujiazui, Shanghai, China in 2006 (Scene II shown in Fig. \ref{fig:3datasets} (b) and (e)). In particular, the data was collected over the urban centre of Shanghai, with multiple curved buildings.  The average density of the point cloud in this region is 1.20 $pts/m^2$. There are five typical curved buildings within these datasets. Building III is very irregular with roofs and surfaces all curved. The building I and II are typical buildings with domical roofs. On the contrary, buildings IV and V are polyhedrons with partially curved roofs or walls.\par 

Furthermore, we run experiments in ‘Roy Thomson Hall’, located in Toronto, Canada (Scene III shown in Fig. \ref{fig:3datasets} (c) and (f)), to compare the proposed modelling method with another roof based approach \cite{zhang2018large} applied to the same building. As the most commonly used regular building reconstruction method, roof based research has achieved both high accuracy and low memory usage. The goal of comparing our proposed method with roof based modelling is to validate that the proposed approach preserves an improved accuracy in curved surface modelling. The LiDAR data ‘Roy Thomson Hall’ was collected with an airborne laser scanner ALTM ORION M (Optech, Canada), and download from the ISPRS LiDAR reconstruction dataset\footnote{http://www2.isprs.org/commissions/comm3/wg4/tests.html}.  Data was acquired in February 2009, taken at the flying altitude of 650 m. The mean average point density is approximately 6.0 $pts/m^2$.\par 

{More tests are conducted on scenarios with dense (Dublin \cite{laefer20172015}) and sparse (Vancouver \footnote{https://opendata.vancouver.ca/explore/dataset/lidar-2013/information/}) points dataset. In the Dublin dataset. The average density of Dublin dataset is 200 $pts/m^2$ while the average  density of Vancouver is 3 $pts/m^2$. Quantitative experiments are conducted on the the two datasets to visualize the point to plane distance error.}\par

\subsection{Building primitive division with {MPA}}

We demonstrate that the proposed modified {Procrustean} analysis provides improved results for primitive contour clustering. We begin by generating a synthetic dataset to test with our algorithm. For a quantitative comparison, no strict criteria are placed to test how contour clusters should be classified into several primitives. However, the synthetic dataset from the assembly of the deformed primitives is an ideal way for validation. Hence, we perform several synthetic experiments from which we derive a qualitative understanding of whether the proposed approach is capable of {managing} complex situations.\par 
\begin{figure}[]
	\centering
	\includegraphics[width=0.5\textwidth]{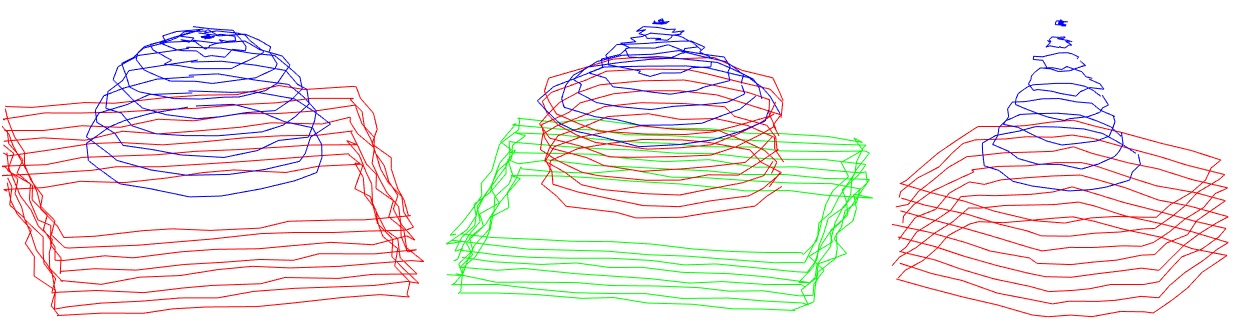}
	\caption{Typical synthetic dataset for testing efficiency of {MPA.}}
	\label{fig:procrustean_test}
\end{figure}
We randomly assemble a hemisphere, cone, cylinder, and polyhedron (random corners) in one cluster and arbitrarily deform the assembled building using the deformation field defined in Section \ref{model_EDfield}. The height of the generated building is 20-30 m with a contour gap of 1 m. In the simulation tests, we consider the scenarios in which contours are stacked upon each other. To make the scenario more challenging, we add {noise} to the corners of each contour using Gaussian distributed noise with a mean of 0 and the sigma of 0.4. Fig. \ref{fig:procrustean_test} shows a typical assembly of primitives obtained from one realization of a Monte Carlo trial of noise and deformation. In this test, we perform 30 Monte Carlo simulations and all {Procrustean} results are compared with ground truth labels. If one contour does not match the label, we assume that the trial fails. The size of the building ranges within a $10 \times 10$ area and a height of less than 20 m. Table \ref{Table:Monte_Carlo_Simulation} presents the accuracy relating to different noise levels. The sensitivity to noise of both methods is similar in small deformations. However, as already discussed, the deformation in curved walls does not ensure a correct orthogonal transformation in the original {Procrustean} analysis. The proposed affine based {Procrustean} analysis significantly addresses the affine transformation and achieves satisfying results for heavy deformations, as shown in Table \ref{Table:Monte_Carlo_Simulation}. Moreover, thanks to the linear formulation of an affine ‘rotation’ in the modified {Procrustean} analysis, the energy function converges much faster than the conventional Euler angle rotation formulation in the original {Procrustean} analysis.\par

\subsection{Storage cost comparison}
The main appeal of the proposed primitive based modelling over previous mesh reconstructions \cite{song2015extraction,wu2017graph} is the storage cost of the model. \cite{song2015extraction} points out that a free form curved surface requires a large amount of storage corners and connections which are extremely difficult to compress due to the irregular data form. Compared with conventional regular roof based reconstructions \cite{sampath2010segmentation,haala2010update,lin2013semantic,xu2015investigation}, the storage cost is equivalent to hundreds of regular buildings. Our strategy of primitive modelling plus deformation refinement greatly reduces the memory storage to several percents of the original mesh storage format.\par 

Recall that the primitive {is consisted of} four domains: the primitive type, the associated parameters in canonical space $\mathbf{\zeta}_i$, the primitive rotation $\mathbf{R}_i$ and the primitive translation $\mathbf{T}_i$. Unlike the mesh model, which is dominated by a large number of patches, this primitive model requires a minimal memory cost. Apart from primitive information, the deformation has become a significant proportion of the overall cost. However, compared with the mesh fitting of the curved surface with small patches, our top-down deformation
field has a great advantage over the bottom-up method. Inherently defined to generate curvature, the deformation field is composed of several hundred to several thousand deformation nodes, each containing 12 parameters ($\mathbf{A}_j$ and $\mathbf{t}_j$). The two primitives are shown in the first two blocks, while the last block records all nodes. For a model like this, a simple mesh presentation is at least 40 segments in each level, and 20 levels overall. According to the definition of Standard Template Library (STL) model or Stereolithography (SL)  models, a small patch is defined by the index of each corner. Therefore, this 800 patch model costs 123 kb. However, a typical deformation graph with a mesh model of 200 nodes costs 25 kb. The patch size increases significantly as the curvature or complexity of the model increases. The ED graph, however, remains stable in practice. According to \cite{newcombe2015dynamicfusion,dou2017motion2fusion,dou2016fusion4d}, 2000 nodes is perfectly adequate for expressing deformations, even in the human body. After testing, we find the upper limit size of the proposed model to be 267 kb ({with 2000 ED nodes upmost}) for fine scale modelling. While it is reported in \cite{song2015extraction} that the reconstructed model ranges from several megabytes (mb) to more than 10 mb.
\par 
\begin{table*}[!h]
	\caption{The percentage (\%) of correct trial. The comparison of contour splitting between {Procrustean} analysis (PA) and modified {Procrustean} analysis (MPA). We consider the situation of different arbitrary deformation and different perturbations exerted (Gaussian noises with 0 mean and $\sigma$ variance) on contour lines. }
	\label{Table:Monte_Carlo_Simulation}
	\begin{center}
		\begin{tabular}{p{0.8cm}|p{3.6cm}|p{3.8cm}|p{3.6cm}}
			ine
			$\sigma$& Deformation Small (PA/MPA) & Deformation Medium (PA/MPA) & Deformation Heav y(PA/MPA) \\
			ine
			1.00  &100.00/100.00 &93.33/100.00 & 90.00/100.00 \\
			
			2.00  &100.00/100.00 &96.67/96.67  & 90.00/96.67 \\
		
			3.00  &96.67/96.67 &93.33/96.67 & 86.66/93.33 \\
		
			4.00  &96.67/100.00 &90.00/96.67 & 83.33/90.00\\
		
			5.00 &93.33/93.33 &86.66/90.00 & 80.00/90.00 \\
			ine
		\end{tabular}
	\end{center}
	\label{Table:monte_carlo_results}
\end{table*}
\begin{figure}[]
	\centering
	\includegraphics[width=0.5\textwidth]{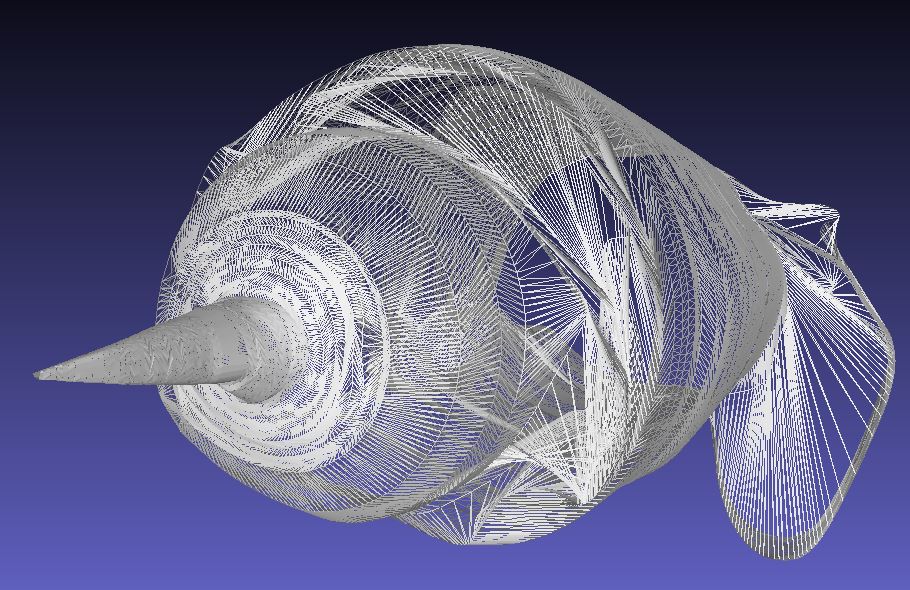}
	\caption{An example of patch complexity in a mesh model.}
	\label{fig:mesh_complexity}
\end{figure}

\begin{figure*}[!h]
	\centering
	\subfloat[]{
		\begin{minipage}[]{0.3\textwidth}
			\centering
			\includegraphics[width=1\linewidth]{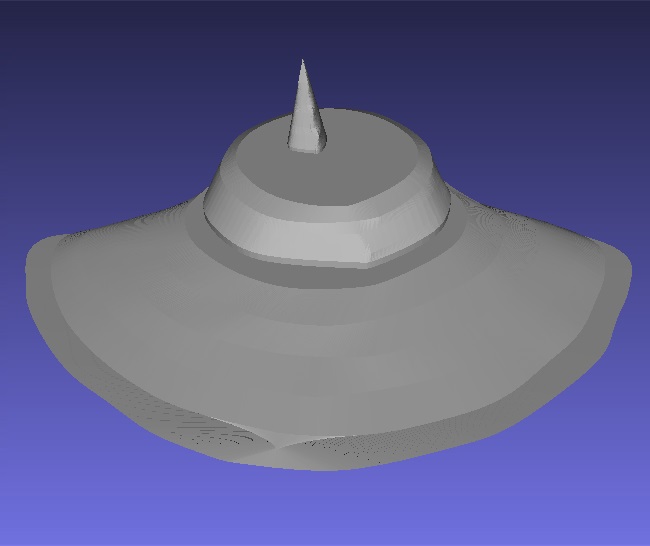}
		\end{minipage}
	}\/
	\subfloat[]{
		\begin{minipage}[]{0.31\textwidth}
			\centering
			\includegraphics[width=1\linewidth]{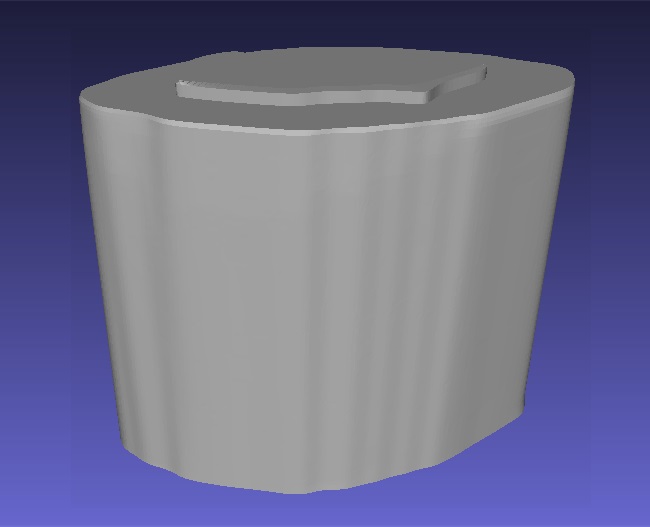}
		\end{minipage}
	}\/
	\subfloat[]{
		\begin{minipage}[]{0.3\textwidth}
			\centering
			\includegraphics[width=1\linewidth]{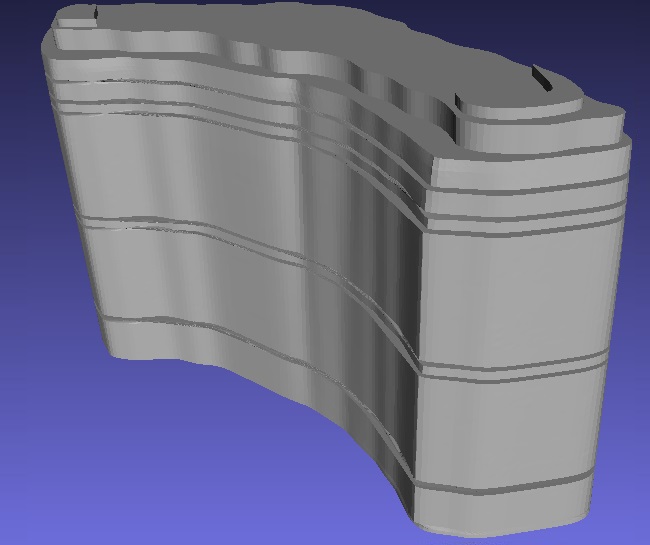}
		\end{minipage}
	}\/
	\caption{Reconstructed building of Sandiago city center (Fig. \ref{fig:3datasets} (a)). (a - c) are the three curved surface buildings.}
	\label{fig:reconstruction_sandiego}
\end{figure*}

\begin{figure*}[!h]
	\centering
	\subfloat[]{
		\begin{minipage}[]{0.2\textwidth}
			\centering
			\includegraphics[width=1\linewidth]{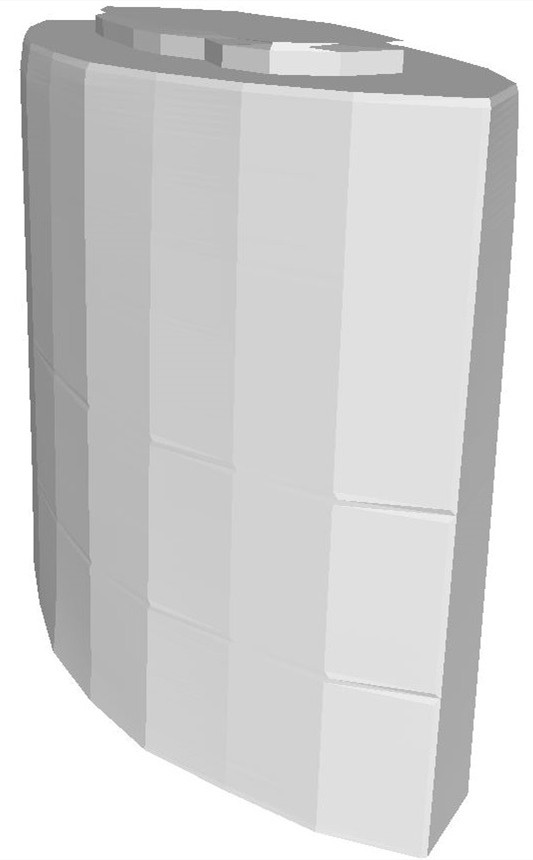}
		\end{minipage}
	}
	\subfloat[]{
		\begin{minipage}[]{0.2\textwidth}
			\centering
			\includegraphics[width=1\linewidth]{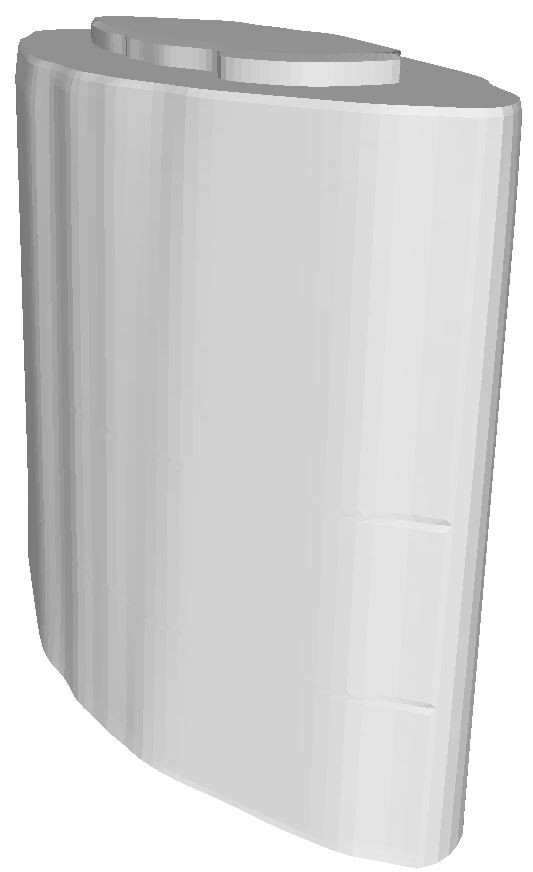}
		\end{minipage}
	}\/
	\subfloat[]{
		\begin{minipage}[]{0.2\textwidth}
			\centering
			\includegraphics[width=1\linewidth]{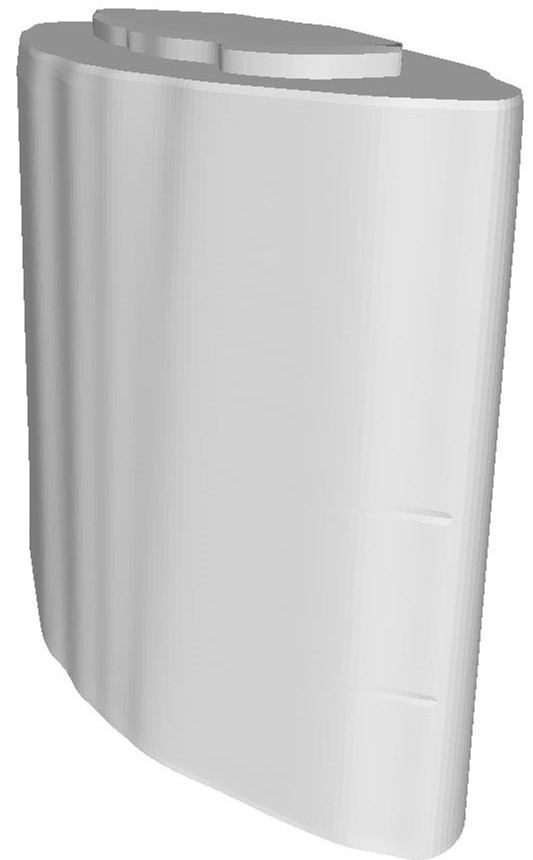}
		\end{minipage}
	}\/
	\subfloat[]{
		\begin{minipage}[]{0.2\textwidth}
			\centering
			\includegraphics[width=1\linewidth]{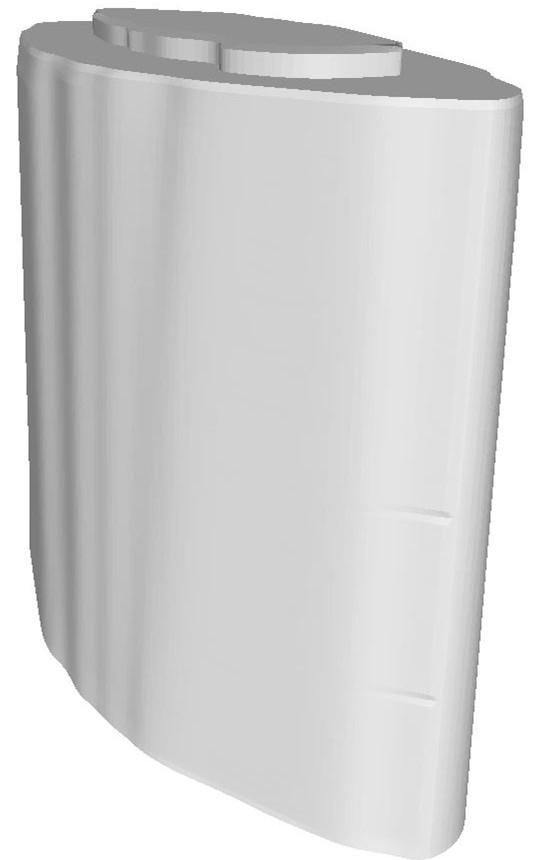}
		\end{minipage}
	}\/
	\caption{(a) is original model without ED graph refinements. (b-d) are the node size with 128, 256 and 512.}
	\label{fig:EDgraphCompare}
\end{figure*}

In essence, the mesh contains serious information redundancy. After fitting the extracted point cloud \cite{song2015extraction}, mesh fitting not only preserves the topology, but also a large amount of noise that should be filtered. The proposed top-down primitive based method is robust to noise. Fig. \ref{fig:mesh_complexity} shows the mesh of previous mesh based reconstruction \cite{song2015extraction}. There are large sum redundant patches on the surface of the model which significantly wastes the storage. The patches evenly scattered on the surface of the building even on the flat surface. By eliminating these unwanted patches, data storage can be greatly saved.\par

Table \ref{table:comparison_of_memory} shows the memory cost of the proposed work and previous research \cite{song2015extraction}. It can be seen that as the model curves more and more, the size of the patch used to express the model increases dramatically.  The proposed primitive based model, however, preserves the curvature without sacrificing memory loss. Moreover, the mesh based approach uniformly increases the resolution of all meshes in order to preserve local details in several regions. Even by ignoring details, the curved surface itself is memory costing. The proposed method, however, exploits the ED graph and primitives in order to present the curved surface in a memory efficient way. Thus, it exhibits more potential in complex models. \par


\begin{table}[ht]
	\caption{{Memory cost comparisons between primitive and mesh based reconstructions . From the reconstructed buildings in Scene II, shown in Fig.6 (b)}. The storage cost is represented in kb.} 
	\centering 
	\begin{tabular}{c|c|c|c} 
		ine 
		$\delta$& Free form mesh model & \textbf{The proposed method}  \\ [1.0ex] 
		ine 
		I & 30375 & 622 \\ 
		II & 7352 & 435\\
		III & 6775 & 1235 \\
		IV & 19268 & 347 \\
		V & 5838 & 257 \\
		[1.0ex] 
		ine 
	\end{tabular}
	\label{table:comparison_of_memory} 
\end{table}

\subsection{3D reconstruction comparisons}
ED nodes are generated by downsampling the original model described in Section \ref{model_EDfield}. The weights for the optimizing energy function defined in Eq. (\ref{energyfunction}) 
are chosen as $w_{rot}=1, w_{reg}=10, w_{data}=100$ similar to \cite{sumner2007embedded}.  \par

\subsubsection{Scene I}
The results of three building reconstructions from San Diego are illustrated in Fig. \ref{fig:reconstruction_sandiego}. Building I (Fig. \ref{fig:reconstruction_sandiego}(a)) is a typical primitive based model assembled by two cones (top and bottom) and a sphere located in the middle.  The proposed {MPA} method accurately identifies the classification. This model is the most ideal case because all buildings are fitted with three simple primitives, thus recording very few parameters. We also choose a small ED (80 nodes) graph for building simplification. The memory usage for this model is 10 kb. Readers may realize that the structure of the building is well preserved in reference to aerial images (Fig. \ref{fig:3datasets}(a)). However, the deformed surface is not very smooth, probably due to a coarse deformation field from a small number of ED nodes.\par 

Buildings II and III (Fig. \ref{fig:reconstruction_sandiego}(b) and (c)) are classified as polyhedrons after contour clustering as the error generated for the contour cluster fitting with the primitives is larger than the thresholds.  However, thanks to the ED graph, a simplified polyhedron with approximately 20 corners is sufficient for the model. {The details of the curved surface are embodied in the deformation field defined by the ED graph. Fig. {\ref{fig:EDgraphCompare}} shows a typical comparison model with different parameters of the ED graph. It shows significant difference between (a-c) while the surfaces of (c) and (d) are similar. Thus, the selection is empirical and is depend on the density of the point cloud.} {Fig. {\ref{fig:EDgraphCompare}} and Table \ref{table:comparison_of_memory} validate our claim that the primitive based curved building reconstruction achieve similar results while keeps low memory cost.} \par 

\begin{table}[]
\caption{Comparisons of the free form and primitive based reconstruction, indexed as the average approximate vertical distances from the segmented point cloud to the reconstructed models (m). The results are in similar sequence of Fig. \ref{fig:reconstruction_compare_withOPTIK}.}
\begin{tabular}{l|l|l|l|l|l}
ine
                & I      & II      & III      & IV      & V      \\ ine
free form       & 0.2749 & 0.2777 & 0.3312 & 0.3495 & 0.1745 \\
Primitive based & 0.4262 & 0.3749 & 0.6338 & 0.7016 & 0.3411 \\ ine
\end{tabular}
\label{Table_quantitative_compare}
\end{table}
{Similar to {\cite{song2015extraction}}, we conduct a quantitative accuracy comparison test. Since it is difficult to directly measure the point to plane distance, we approximate it by transforming the model into very dense point
clouds, and the average accuracy is measured by the point to point distance from the raw data to the model. The results are represented in Table. {\ref{Table_quantitative_compare}}. Similar to Fig. {\ref{fig:reconstruction_compare_withOPTIK}}, the primitive based method is inferior to the free form methods but the structure is well preserved.}\par

\subsubsection{Scene II}
\begin{figure*}[!htpb]
	\centering
	\subfloat[]{
		\begin{minipage}[]{0.23\textwidth}
			\centering
			\includegraphics[width=1\linewidth]{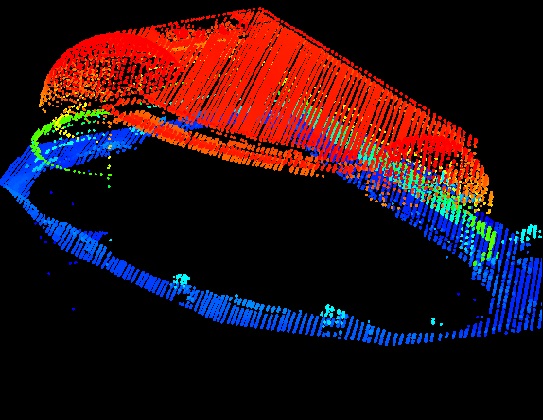}
		\end{minipage}
	}\/
	\subfloat[]{
		\begin{minipage}[]{0.23\textwidth}
			\centering
			\includegraphics[width=1\linewidth]{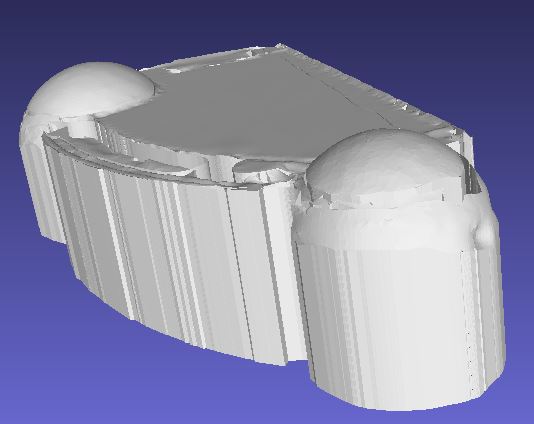}
		\end{minipage}
	}\/
	\subfloat[]{
		\begin{minipage}[]{0.23\textwidth}
			\centering
			\includegraphics[width=1\linewidth]{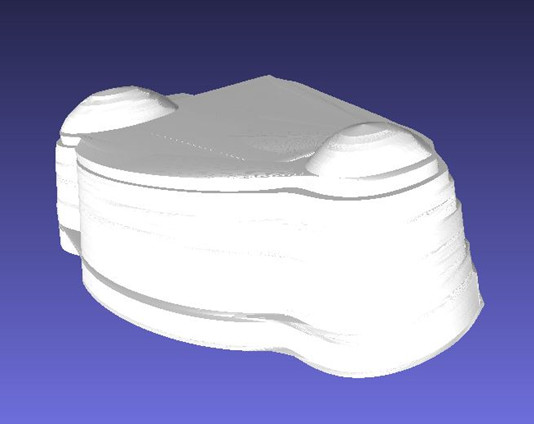}
		\end{minipage}
	}\/ \\
	\subfloat[]{
		\begin{minipage}[]{0.23\textwidth}
			\centering
			\includegraphics[width=1\linewidth]{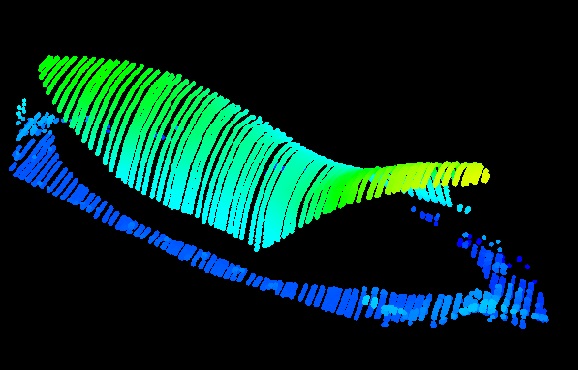}
		\end{minipage}
	}\/
	\subfloat[]{
		\begin{minipage}[]{0.23\textwidth}
			\centering
			\includegraphics[width=1\linewidth]{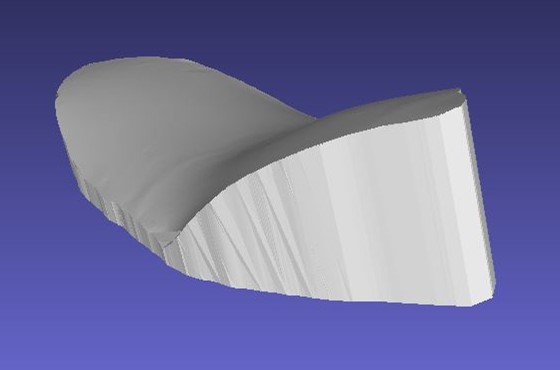}
		\end{minipage}
	}\/
	\subfloat[]{
		\begin{minipage}[]{0.23\textwidth}
			\centering
			\includegraphics[width=1\linewidth]{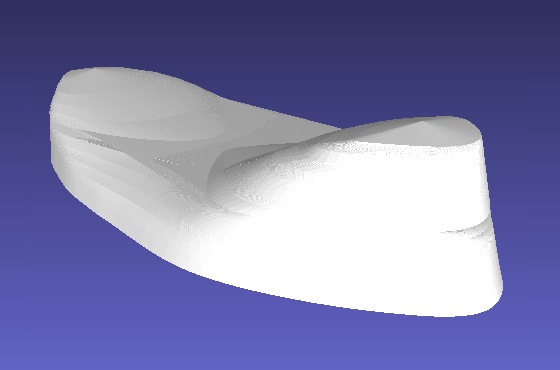}
		\end{minipage}
	}\/\\
	\subfloat[]{
		\begin{minipage}[]{0.23\textwidth}
			\centering
			\includegraphics[width=1\linewidth]{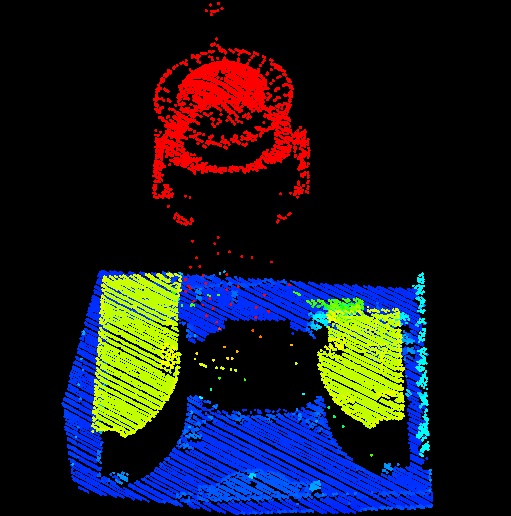}
		\end{minipage}
	}\/
	\subfloat[]{
		\begin{minipage}[]{0.23\textwidth}
			\centering
			\includegraphics[width=1\linewidth]{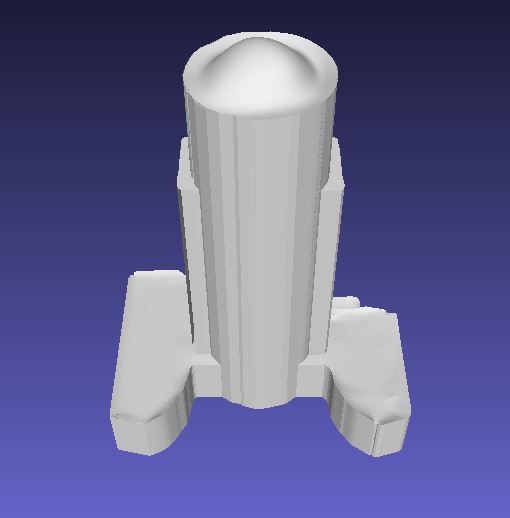}
		\end{minipage}
	}\/
	\subfloat[]{
		\begin{minipage}[]{0.23\textwidth}
			\centering
			\includegraphics[width=1\linewidth]{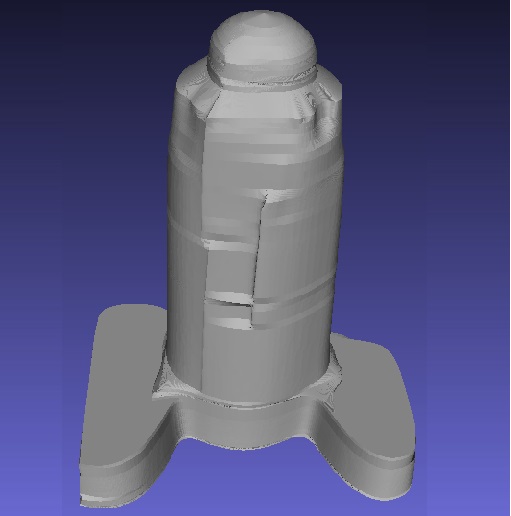}
		\end{minipage}
	}\/\\
	\subfloat[]{
		\begin{minipage}[]{0.21\textwidth}
			\centering
			\includegraphics[width=1\linewidth]{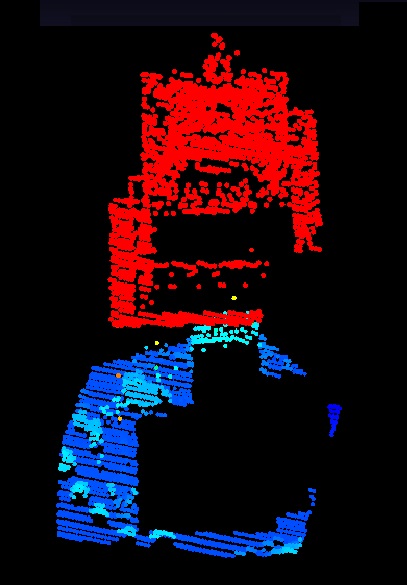}
		\end{minipage}
	}\/
	\subfloat[]{
		\begin{minipage}[]{0.21\textwidth}
			\centering
			\includegraphics[width=1\linewidth]{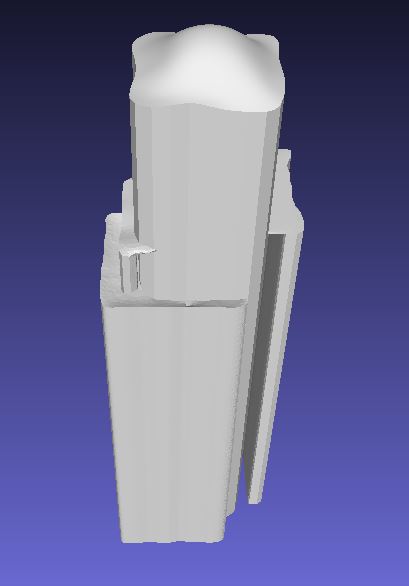}
		\end{minipage}
	}\/
	\subfloat[]{
		\begin{minipage}[]{0.21\textwidth}
			\centering
			\includegraphics[width=\linewidth]{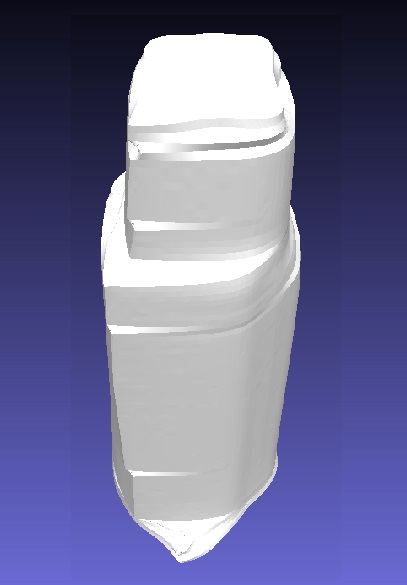}
		\end{minipage}
	}\/\\
	\subfloat[]{
		\begin{minipage}[]{0.3\textwidth}
			\centering
			\includegraphics[width=1\linewidth]{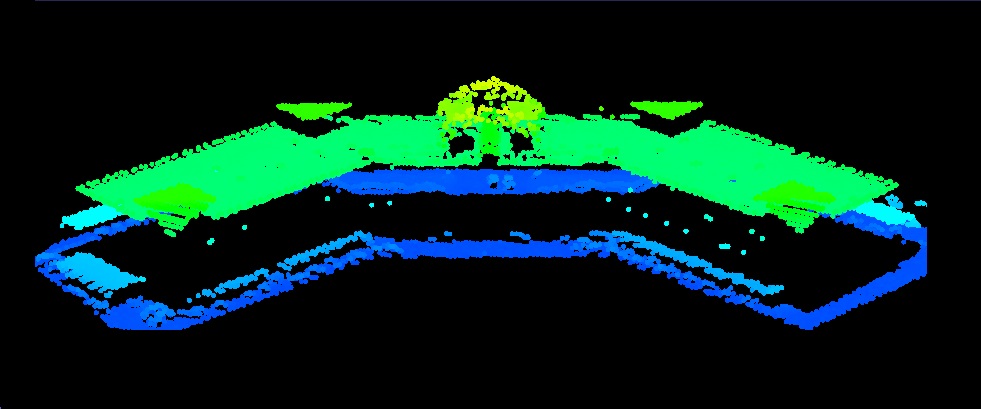}
		\end{minipage}
	}\/
	\subfloat[]{
		\begin{minipage}[]{0.3\textwidth}
			\centering
			\includegraphics[width=1\linewidth]{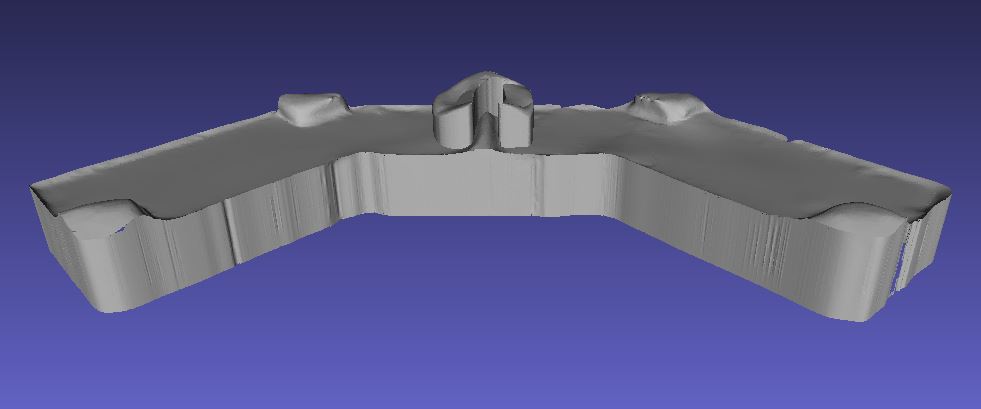}
		\end{minipage}
	}\/
	\subfloat[]{
		\begin{minipage}[]{0.3\textwidth}
			\centering
			\includegraphics[width=1\linewidth]{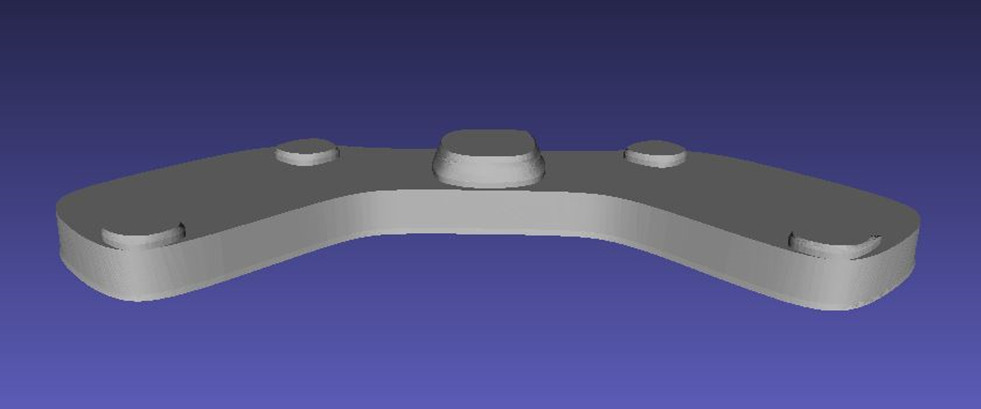}
		\end{minipage}
	}\/
	\caption{Qualitative comparisons of proposed method and \cite{song2015extraction}. The left most column shows original point cloud. The middle shows models in \cite{song2015extraction} while the last column gives the results created by the proposed framework.}
	\label{fig:reconstruction_compare_withOPTIK}
\end{figure*}

We present a model quality assessment as well as memory usage comparisons between the proposed primitive based approach and previous mesh based reconstructions \cite{song2015extraction}. Fig. \ref{fig:reconstruction_compare_withOPTIK} illustrates the reconstruction from two different approaches. In Fig. \ref{fig:reconstruction_compare_withOPTIK}, models (c), (f) and (o) preserve a better smoothness compared to the original model. The structure of the uniformly reconstructed model lacks some local details, but preserves superior global structures. Note that the structures generated by our work can be further processed with manually editing.\par 
For buildings (i) and (l) in Fig. \ref{fig:reconstruction_compare_withOPTIK}, the disorder on the wall is mainly attributed to the disorder of joints within consecutive contour clusters. After analyzing the original point cloud and the reconstructed wall, we realize that the LiDAR points on the facade are missing due to the scanning angle of the sensor. The missing information makes the generated contours unpredictable. As information regarding the contours is not enough, the contour based method alone is not able to generate improvement details and thus more priori information should be integrated for facade predictions.\par

\begin{figure*}[!h]
	\centering
	\subfloat[]{
		\begin{minipage}[]{0.23\textwidth}
			\centering
			\includegraphics[width=1\linewidth]{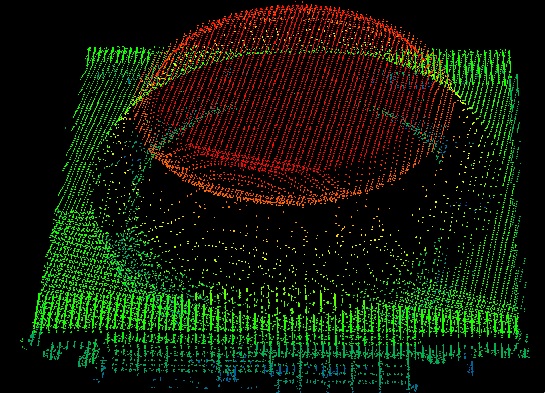}
		\end{minipage}
	}\/
	\subfloat[]{
		\begin{minipage}[]{0.23\textwidth}
			\centering
			\includegraphics[width=1\linewidth]{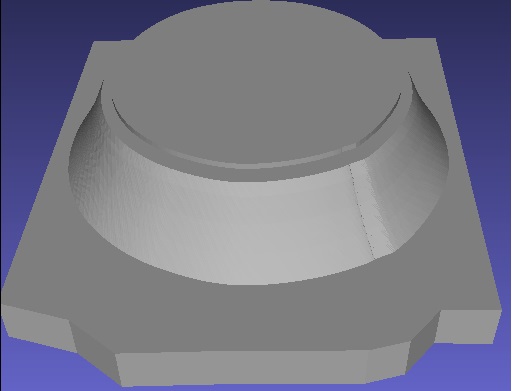}
		\end{minipage}
	}\/
	\subfloat[]{
		\begin{minipage}[]{0.23\textwidth}
			\centering
			\includegraphics[width=1\linewidth]{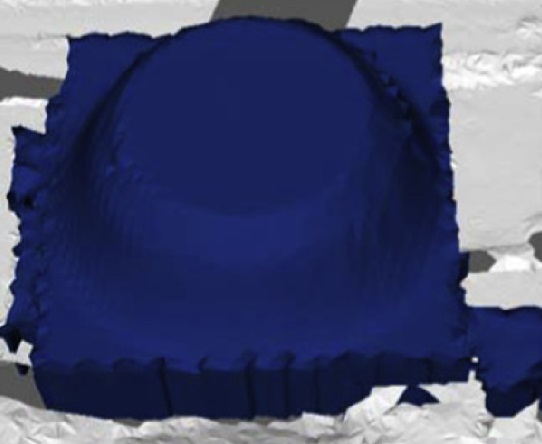}
		\end{minipage}
	}
	\subfloat[]{
		\begin{minipage}[]{0.23\textwidth}
			\centering
			\includegraphics[width=1\linewidth]{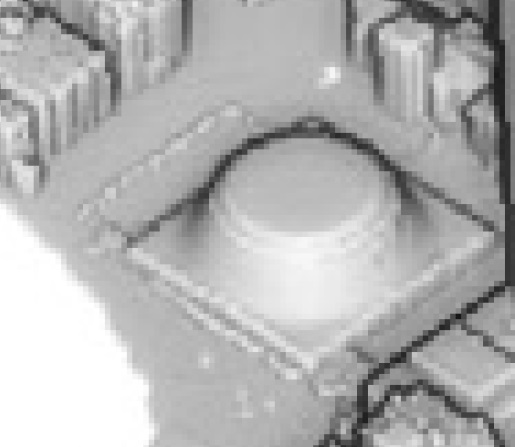}
		\end{minipage}
	}
	\caption{ The comparisons between our method and \cite{zhang2018large}. (a) is the original point cloud. (b) is the 3D reconstruction by our method. (c) is the result from \cite{zhang2018large}. (d) is another result from \cite{zhang2018large}.}
	\label{fig:reconstruction_comparison}
\end{figure*}

\begin{figure*}[!h]
	\centering
	\subfloat[]{
		\begin{minipage}[]{0.23\textwidth}
			\centering
			\includegraphics[width=1\linewidth]{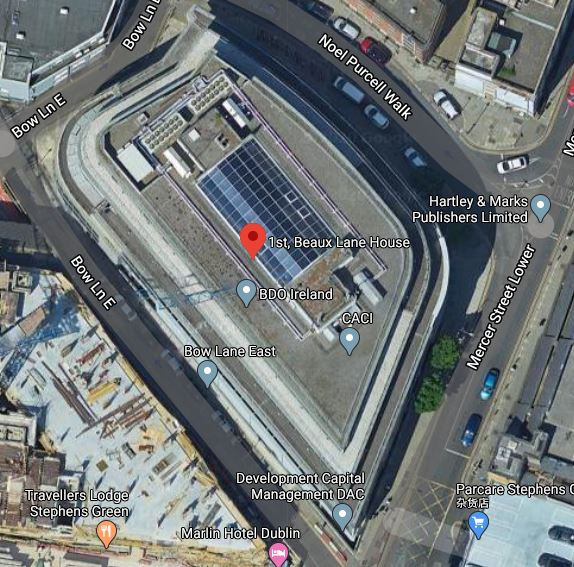}
		\end{minipage}
	}
	\subfloat[]{
		\begin{minipage}[]{0.23\textwidth}
			\centering
			\includegraphics[width=1\linewidth]{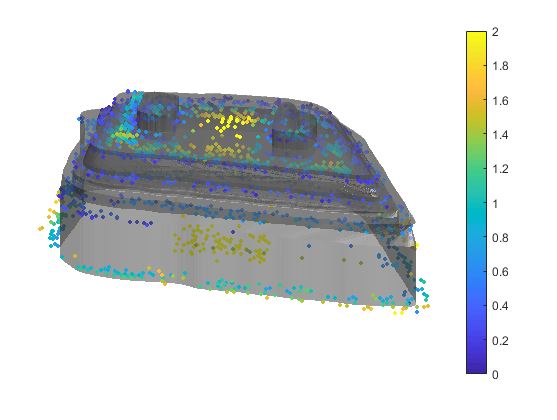}
		\end{minipage}
	}\/
	\subfloat[]{
		\begin{minipage}[]{0.23\textwidth}
			\centering
			\includegraphics[width=1\linewidth]{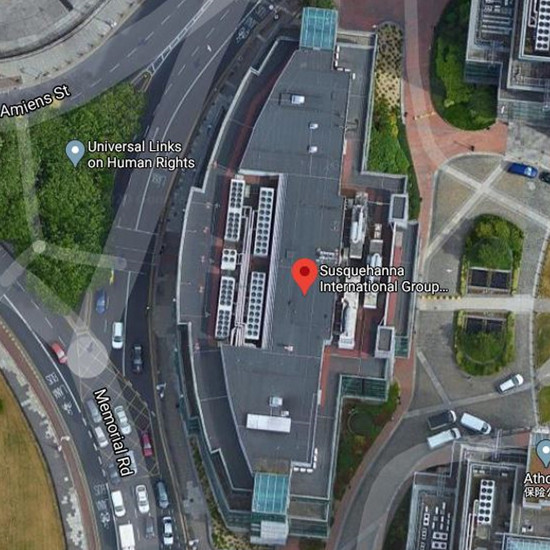}
		\end{minipage}
	}
	\subfloat[]{
		\begin{minipage}[]{0.23\textwidth}
			\centering
			\includegraphics[width=1\linewidth]{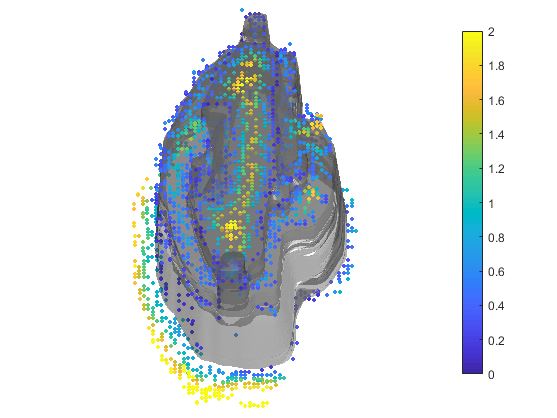}
		\end{minipage}
	}

	
	\caption{ The 3D reconstructed models in the Dublin dataset. (a) and (c) are the satellite images. (b) and (d) are the corresponding recovered buildings with the point cloud. }
	\label{fig:reconstruction_Dublin}
\end{figure*}

\begin{figure*}[!h]
	\centering
	\subfloat[]{
		\begin{minipage}[]{0.45\textwidth}
			\centering
			\includegraphics[width=1\linewidth]{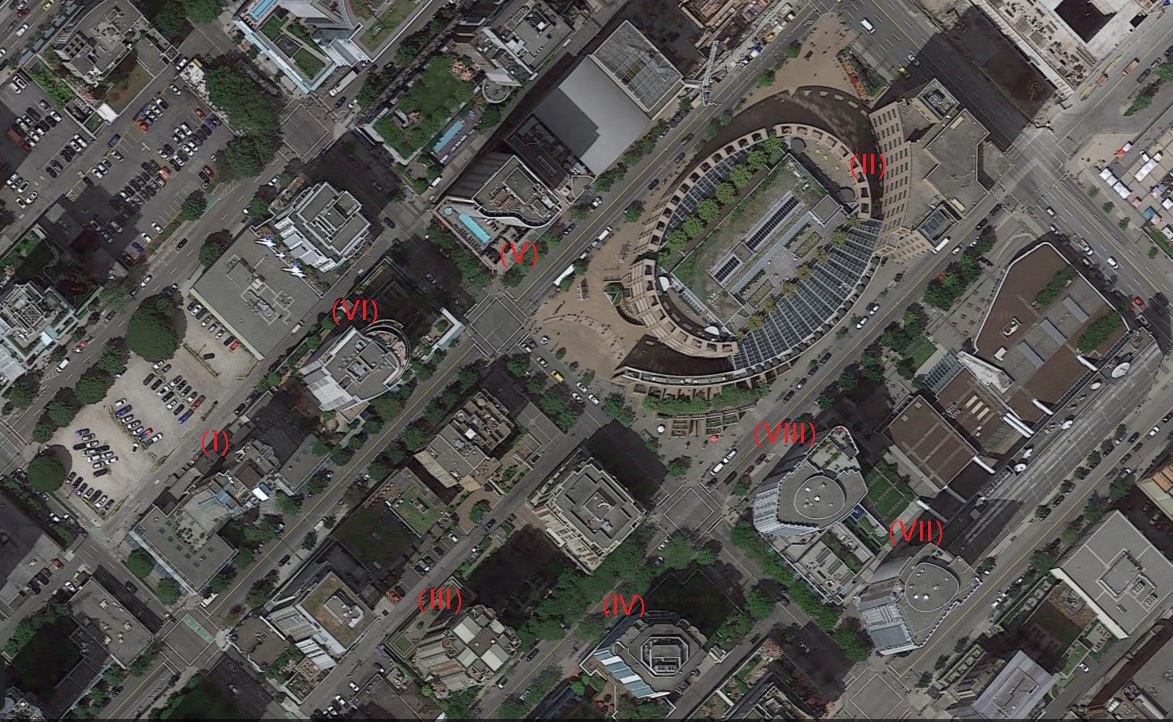}
		\end{minipage}
	}\/
	\subfloat[]{
		\begin{minipage}[]{0.45\textwidth}
			\centering
			\includegraphics[width=1\linewidth]{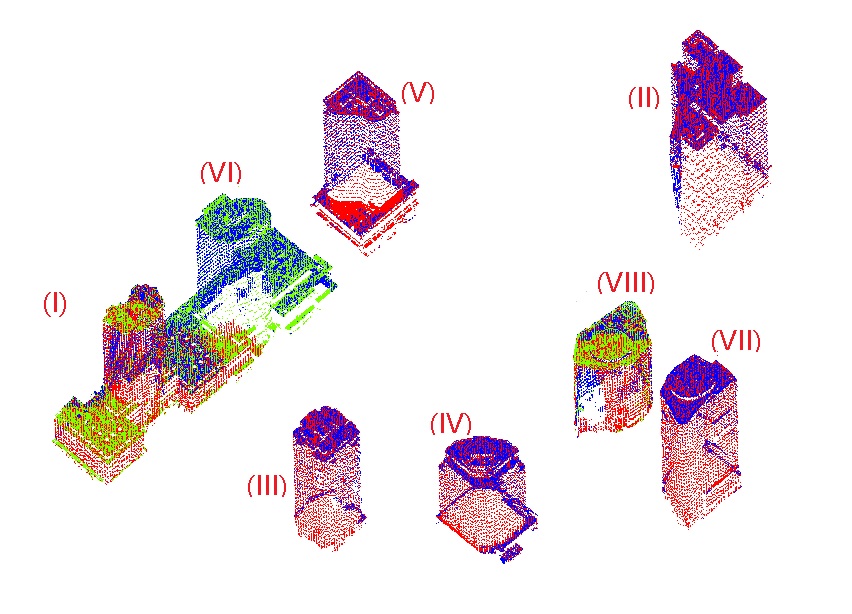}
		\end{minipage}
	}\/
	\subfloat[Building I]{
		\begin{minipage}[]{0.23\textwidth}
			\centering
			\includegraphics[width=1\linewidth]{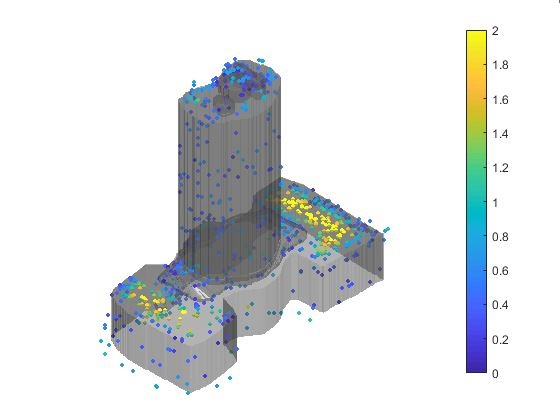}
		\end{minipage}
	}\/
	\subfloat[Building II]{
		\begin{minipage}[]{0.23\textwidth}
			\centering
			\includegraphics[width=1\linewidth]{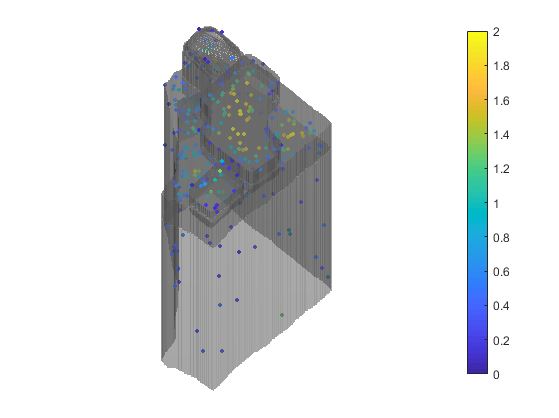}
		\end{minipage}
	}\/
	\subfloat[Building III]{
		\begin{minipage}[]{0.23\textwidth}
			\centering
			\includegraphics[width=1\linewidth]{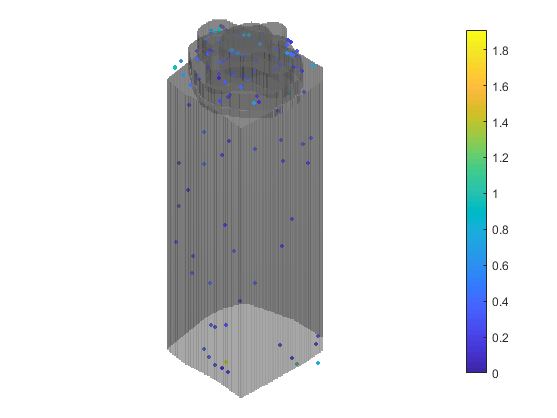}
		\end{minipage}
	}\/
	\subfloat[Building IV]{
		\begin{minipage}[]{0.23\textwidth}
			\centering
			\includegraphics[width=1\linewidth]{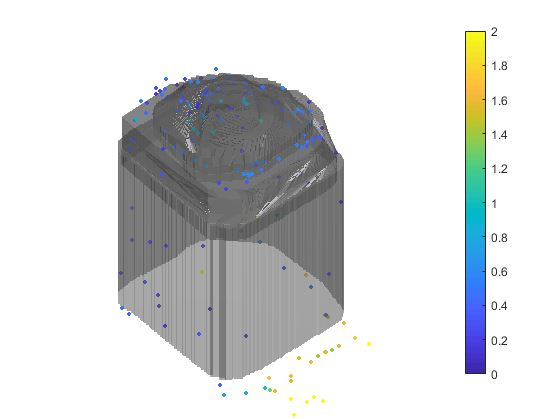}
		\end{minipage}
	}\/
	\subfloat[Building V]{
		\begin{minipage}[]{0.23\textwidth}
			\centering
			\includegraphics[width=1\linewidth]{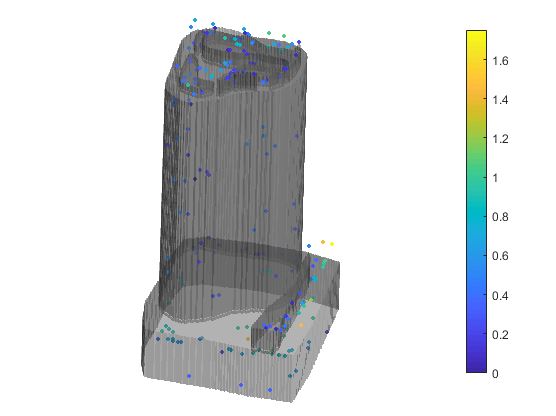}
		\end{minipage}
	}\/
	\subfloat[Building VI]{
		\begin{minipage}[]{0.23\textwidth}
			\centering
			\includegraphics[width=1\linewidth]{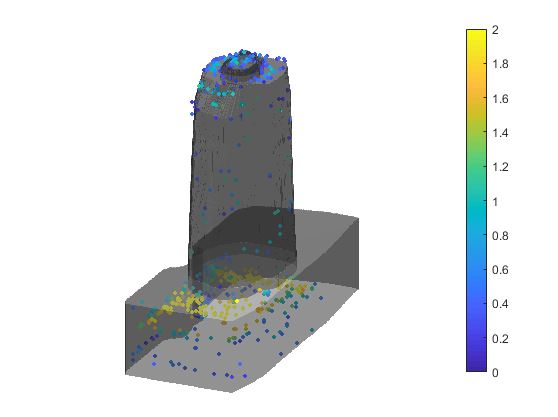}
		\end{minipage}
	}\/
	\subfloat[Building VII]{
		\begin{minipage}[]{0.23\textwidth}
			\centering
			\includegraphics[width=1\linewidth]{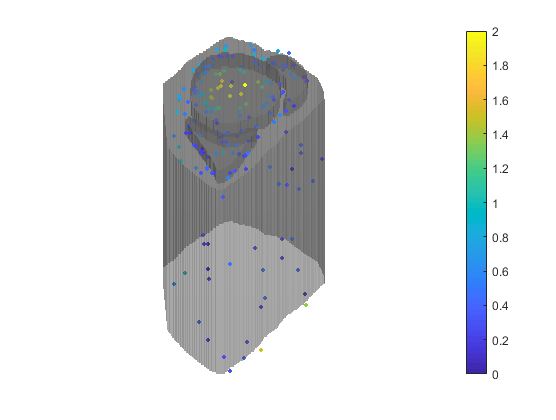}
		\end{minipage}
	}\/
	\subfloat[Building VIII]{
		\begin{minipage}[]{0.23\textwidth}
			\centering
			\includegraphics[width=1\linewidth]{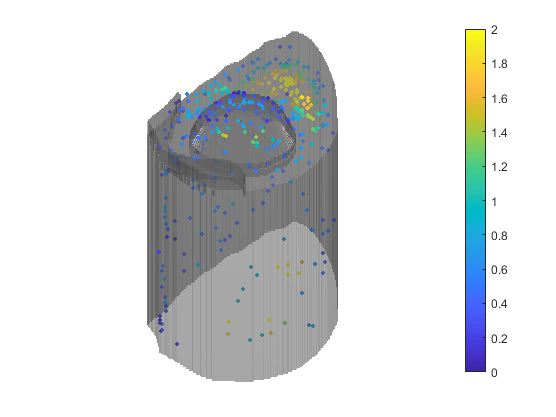}
		\end{minipage}
	}\/
	
	\caption{ The 3D reconstructed models in the Dublin dataset. (a) and (d) are the satellite images. (b) and (e) are the related point cloud. (c) and (f) are the reconstructed }
	\label{fig:reconstruction_Vancourver}
\end{figure*}
\subsubsection{Scene III}
Fig. \ref{fig:reconstruction_comparison} demonstrates qualitative comparisons of the proposed approach and the reconstruction results from state-of-the-art \cite{zhang2018large}. Fig. \ref{fig:3datasets} demonstrates that this building is a double roof building. Our work successfully divides building contour cluster with modified {Procrustean} analysis and clearly shows the roof in Fig. \ref{fig:reconstruction_comparison}(b). The reconstruction in \cite{zhang2018large}, shown in Figs. 12 (c) and (d), fails to preserve the structure of the roofing and presents the building in a coarse form, similar to a rough digital elevation model. As previous work \cite{song2015extraction,he20163d,wu2017graph,gao2017automatic,gilani2018segmentation} indicates, curved surfaces are not suitable to be presented as conventional polyhedron roof types. Therefore, the reconstruction in \cite{zhang2018large} greatly suffers from low-quality free form surface modelling.\par 

\subsubsection{Scene IV and V}
{Fig. \ref{fig:reconstruction_Dublin} demonstrates the 3D reconstructed building overlayed with the original dense point cloud. The point cloud is represented in the form of colored errors showing the point to plane distance. Results demonstrate the reconstruction preserves the large scale structure like walls and part of the roof. However, the fine scale structure on the roof is not recovered. This is mainly due to the bad contour generation. The results indicate that the fine scaled details from the very dense point cloud can hardly be preserved. On the contrary, Fig. \ref{fig:reconstruction_Vancourver} shows a similar quantitative experiment on the Vancouver dataset which is in the form of sparse point cloud. The result shows the coarse scale is preserved well with the proposed method. The contours of the buildings show that the contours of those small components are often ignored and filtered because the area is small. The building contour extraction step can hardly identify the noise or the small component based on their shape.}

\section{Conclusion}
{A primitive based 3D curved building reconstruction from the Airborne LiDAR is proposed in this paper. The contour clusters is adopted to segment building with curved shape. To adapt the the contour clusterings algorithm in  building primitive segmentation, we propose a modified Procrustean analysis approach.} Moreover, to further improve the quality of coarse buildings based on primitives, we import an ED graph based deformation field for 3D building refinement and polishing.\par 
{The proposed reconstruction process is unsupervised and data-driven, explaining the inherent structure of complex curved surface buildings. By discovering and exploiting the consistent structure of the proposed model, the high-level knowledge based reconstruction maintains data storage at a minimal level. Comparing with previous free form reconstruction method that uses mesh models, the proposed innovative primitives based framework is consistent with polyhedral method, significantly reduces data storage and is suitable for storage data management and processing.} We quantitatively prove the benefits of storage reduction. Three different datasets are employed for validation and comparisons in terms of the quality of the reconstruction and memory usage of the model.\par 
The reconstruction of facades without enough observations is still a challenge due to oblique data acquirement. Small information loss may be repaired with the smoothness of contour lines. For large information loss, it may be recovered by symmetry embedded in most buildings. {Moreover, it remains difficult to achieve fully autonomous workflow because it remains difficult to separate mixed types of objects, curved ones and polyhedrons.} Therefore, further work may focus on the introduction of a priori information for reasonable unobservable facade reconstruction{, and figure out new an approach to identify curved, polyhedral and mix-typed buildings.}


%

%
%
%
%
%

\ifCLASSOPTIONcaptionsoff
  \newpage
\fi



\bibliographystyle{IEEEtran}
\bibliography{reference}
\end{document}